\documentclass[journal]{IEEEtran}
\usepackage{needspace}
\usepackage{amsmath}
\usepackage{tensor}
\usepackage{bm}
\usepackage{amssymb}
\usepackage{graphicx}
\usepackage{caption}
\usepackage{color}
\usepackage{midfloat}
\usepackage{float}
\usepackage{rotating}
\usepackage{multirow}
\usepackage{hhline}
\usepackage{booktabs}
\usepackage{makecell}
\usepackage{longtable}
\usepackage{colortbl}
\usepackage{array}
\usepackage{paralist}
\usepackage{multicol}
\usepackage{multirow}
\usepackage{chngcntr}
\usepackage{inconsolata}
\usepackage{arydshln,leftidx,mathtools}
\usepackage{textcomp}
\usepackage{microtype}
\usepackage{cite}
\usepackage{stfloats}
\usepackage[utf8]{inputenc}
\usepackage[T1]{fontenc}
\usepackage{mathtools}
\usepackage[thinc]{esdiff}
\usepackage{url}
\usepackage{hyperref} 

\usepackage{xcolor}
\ifCLASSINFOpdf

\newcommand{\subparagraph}{}

\usepackage[compact]{titlesec}

\usepackage{microtype}
\usepackage{cite}

\newcommand{\xdownarrow}[1]{%
  {\left\downarrow\vbox to #1{}\right.\kern-\nulldelimiterspace}
}

\begin{document}
\title{Soft Robots Modeling: a Structured Overview}
\author{Costanza Armanini$^{*1}$, Frédéric Boyer$^{2}$, Anup Teejo Mathew$^{1}$,  Christian Duriez$^{3}$ and Federico~Renda$^{1,4}$	
\thanks{* Corresponding author {\tt\small costanza.armanini@ku.ac.ae}}
\thanks{$^{1}$ Department of Mechanical Engineering, Khalifa University of Science and Technology, Abu Dhabi, UAE.}
\thanks{$^{2}$ LS2N Laboratory, Institut Mines Telecom Atlantique, Nantes, France}
\thanks{$^{3}$ Inria, CNRS, Centrale Lille, Team DEFROST, University Lille, Lille, France}
\thanks{$^{4}$ KUCARS, Khalifa University of Science and Technology, Abu Dhabi, UAE.}
}

\maketitle

\begin{abstract}
The robotics community has seen an exponential growth in the level of complexity of the theoretical tools presented for the modeling of soft robotics devices. Different solutions have been presented to overcome the difficulties related to the modeling of soft robots, often leveraging on other scientific disciplines, such as continuum mechanics, computational mechanics and computer graphics. These theoretical and computational foundations are often taken for granted and this leads to an intricate literature that, consequently, has rarely been the subject of a complete review. For the first time, we present here a structured overview of all the approaches proposed so far to model soft robots. The chosen classification, which is based on their theoretical and numerical grounds, allows us to provide a critical analysis about their uses and applicability. This will enable robotics researchers to learn the basics of these modeling techniques and their associated numerical methods, but also to have a critical perspective on their uses.

\vspace{-0.3cm}
\end{abstract}

\section{Introduction}
\noindent The term \textit{soft robot} appeared for the first time in a scientific paper in 2000, describing a McKibben pneumatic artificial muscle \cite{Tondu_2000}, a family of braided pneumatic actuators developed in the 50s to assist polio patients. Even though they were not explicitly called \textit{soft robots}, McKibben actuators probably represent the first example of a robotic device exploiting its compliance to achieve improved performances with respect to their exclusively rigid counterpart. Since then, soft robotics has been one of the fastest growing research community in the last decades. Many different soft robotics devices and actuators have been presented, ranging in almost every possible technological field, from biomedical engineering to aerospace and underwater robotics. The increasing interest in soft robotics is demonstrated by the vast number of review papers that have been published to summarize the employed techniques, the achievements and the future prospects of this promising research field  \cite{Laschi2016}, \cite{Trivedi2008_Rev},  \cite{Rus2015},  \cite{Thuruthel2018}. Among the available reviews on the theoretical modeling components, some have been published for specific groups of approaches, \cite{Webster2010}, \cite{Chirikjian2015}, \cite{Chawla2018}, \cite{Daekyum2021}, \cite{Wang_2021}, some for specific application fields \cite{Burgner_TRO_2015}, \cite{Gilbert2021} and others for specific families of robots \cite{Rao2021}. 

\begin{figure}[ht]
\centering
\vspace{-0.2cm}
\includegraphics[width=0.8\columnwidth]{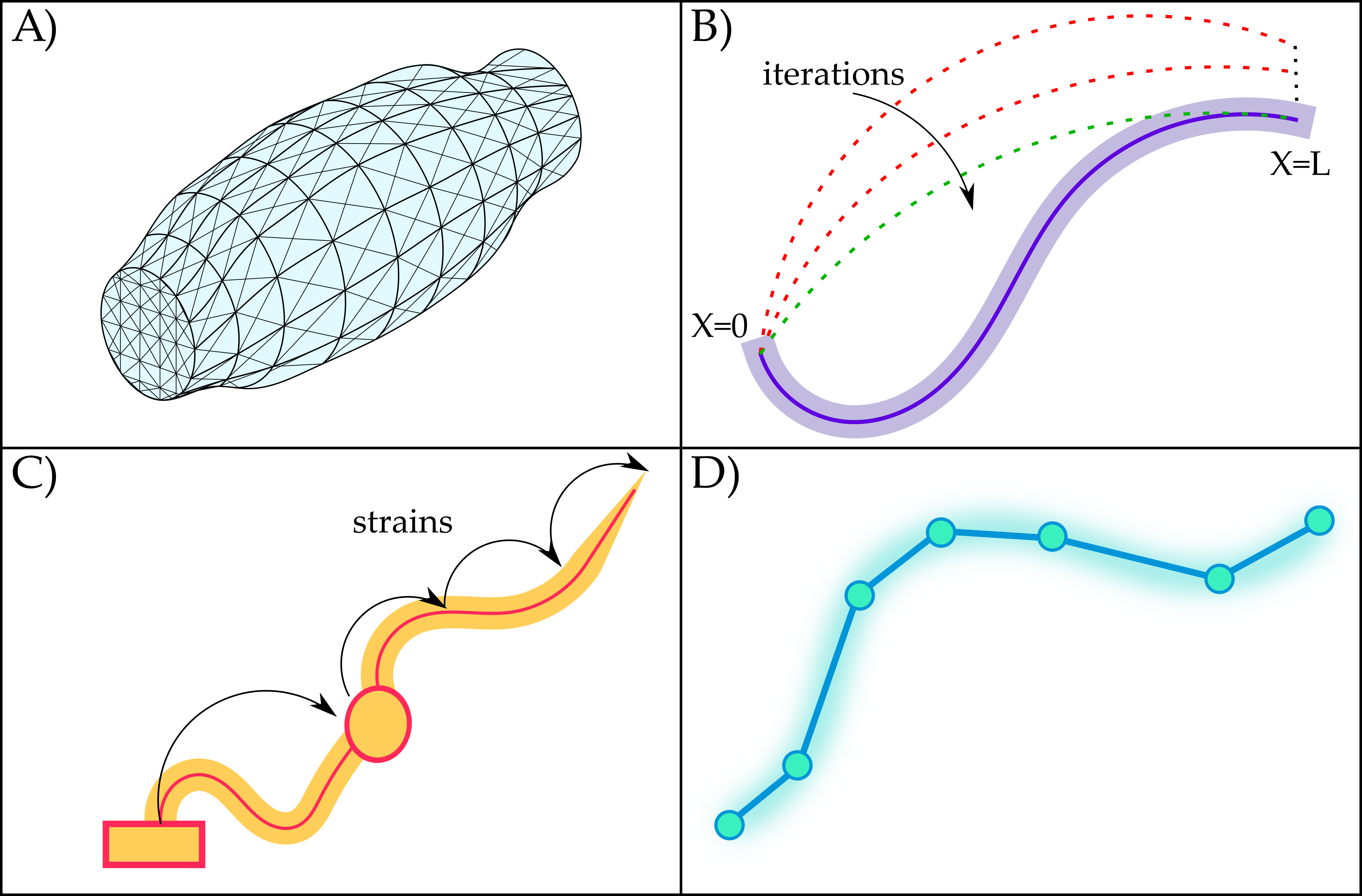}
\caption{\footnotesize Representative examples of the resolution methods that have been distinguished in this review: A) FEM, (Sec. \ref{sec::FEM}), B) Shooting Method (Sec. \ref{sec::Shooting}), C) Relative modal-Ritz reduction (Sec. \ref{sec::RmodalRitz}), D) Pseudo-Rigid approach (Sec. \ref{sec::PseudoRigid}).}
\label{fig:Reductions}
\vspace{-0.2cm}
\end{figure}

\noindent Unlike traditional rigid robots, soft robots are infinite dimensional systems whose time evolution is governed by highly nonlinear partial differential equations, generally not analytically integrable. Based on this observation, the search for a trade-off between accuracy and numerical usability for robotic purposes (such as control and optimization), is undoubtedly the great challenge of soft robot modeling. For the first time, we present here a comprehensive literature review of the techniques that have been presented so far to model soft and continuous robots. Given the enormous quantity and diversity of contributions on this theme in recent years, choices had to be made to classify them. The chosen classification is based on the mathematical techniques themselves, (Fig. \ref{fig:Reductions}), and not on their uses (design, simulation, control) or the designs for which they are used. Although this choice may seem a bit arbitrary at first sight, it has the great advantage of highlighting the structural similarities and differences that gather and/or discriminate all the modeling approaches proposed so far in the field. Moreover, it allows us to go back to the theoretical roots of the different approaches and to give their historical origins, sometimes from other communities. We believe that this will enable robotics researchers to learn the basics of these modeling techniques and their associated numerical methods, and to have a critical perspective on their uses. For example, by revisiting the assumptions on which these models are based when necessary.
This classification will provide a structured overview of all the approaches proposed so far to model soft robots, and will allow us to critically analyze them, according to their uses and the types of systems they are applied to.

\noindent The remaining parts of the paper are organized as follows. In Sec. \ref{sec::Char} some general concepts are drawn and the classification employed in the paper to discuss all the modeling approaches is presented. In Sec. \ref{sec::Phys} we present the formulation of models that are based on continuum mechanics theories, while their numerical resolution is presented in Sec \ref{sec::Num3D} and \ref{sec::Num1D}. In Sec. \ref{sec::Geom}, we present the approaches which rely on the hypothesis that the deformed shape of the robot follows a specific geometrical curve. Proceeding with the level of discretization, in Sec. \ref{sec::Discr} the approaches that arise from an a priori discretization of the soft body are presented, while in Sec. \ref{sec::DataDriven}, data-driven models are discussed. Sec. \ref{sec::Comput} presents the software implementation of the proposed approaches,while in Sec. X, we will critically analyse the applicability of the different approaches. Finally, in Sec. \ref{sec::Conclusions} some conclusions are drawn.

\section{Basic Concepts and Classification}
\label{sec::Char}
Mechanical modelling is a set of mathematical techniques, allowing to represent the evolution of a material system. It is defined by the following key concepts \cite{Bullo2004}, \cite{Murray}:\\
\noindent $\bullet$ \textit{Configuration}: the set of position of the system's particles compatible with the internal (joints) and the external (geometric boundary conditions) kinematic constraints. At any given time, the configuration of a mechanical system defines a subset of the ambient Euclidean space.\\
\noindent $\bullet$ \textit{Generalized Coordinates}: a set of parameters describing any configuration. They are the coordinates of a unique point in an abstract space or "configuration space". They can be \textit{absolute}  when they refer to a fixed inertial frame, or \textit{relative}  when they refer to a moving frame co-varying together with the system.\\
\noindent $\bullet$ \textit{Kinematic map}: it takes as input the generalized coordinates and it returns the configuration of the system. This is also called \textit{forward kinematics}, while the \textit{inverse kinematics} represents exactly the opposite process, i.e. the calculation of the coordinates (typically, the actuation ones) required to obtain a specific configuration of the system.\\
\noindent $\bullet$  \textit{Dynamic principle}: Such as Newton's laws, d'Alembert's principle or Hamilton's principle. Such a principle provides the equations of motion (EoM) of the system governing the temporal evolution of its configurations.\\

\noindent The above characteristics of a mechanical model can be identified relatively easily for a traditional rigid robot. However, this is not the case for soft robots, due to their continuum nature.  
In fact, while traditional rigid robots can be fully represented by some finite discrete set of frames, in soft robotics the robot is a continuum of particles. Let us draw some general considerations on the main steps that are required to obtain a model for a continuum body.
In the Lagrangian description of solid continuum mechanics, the body's \textit{configuration} is parameterized through \textit{positional} fields, which depend both on time and on the material coordinates, i.e. a set of continuous labels $\bm{X}$ that identifies each particle. In particular, $\bm{X}$ can be defined as the coordinates of the particles of the body when it is in a stress-less configuration, called \textit{reference configuration}. The motion of a continuum body is defined by a continuous sequence of configurations along time.
A change in the configuration of a continuum body results in a \textit{displacement}, which usually has two components: a rigid-body displacement and a \textit{deformation}. To describe the internal deformation state of the body, it is necessary to define its strain (time) rate, which is a combination of the gradient of the velocity fields. The definition of the strains needs to be objective, i.e., observer independent. Finally, beyond kinematics, the closed formulation describing the time evolution of a continuum medium is given by: (1) a principle of the dynamics (as before), providing the Partial Differential Equations (PDEs) relating the stress with the acceleration of the particles and the external forces applied inside and across the boundaries of the medium; (2) a set of geometric boundary conditions (BCs); (3) a constitutive law that relates the time evolution of the stress to that of the strains.

\noindent The resulting dynamic equations are, in general, highly nonlinear and characterized by an infinite dimensional configuration space. The soft robotics modeling literature could be viewed as the story of how these extremely complex equations, when applied to soft robots, can be discretized and solved. It should be noted that \textit{discretization} and \textit{reduction} are two distinct concepts that should not be confused. A \textit{discretization} is employed to obtain a numerical solution to the non-linear problem and, for example, the finite differences methods fall in this category. On the other side, a \textit{reduction} consists in the depiction of a basis of functions allowing a proper description of the kinematic fields of the body.
These features are not always clearly expressed in the developing of a theoretical model, or they might be implicitly imposed by the assumptions of the model itself or its numerical resolution.  One of the aims of this work is to shed light on the underline structure of the models proposed in literature using a unified language, which facilitate their understanding and comparison. With this as a guideline, we classified the modeling of soft robots as follows:\\
\noindent 1) \textit{Continuum Mechanics models.} They are characterized by a continuous (infinite-dimensional) configuration space, and on physical considerations about the soft bodies deformations. As such, the models in this category benefit from a physically rigorous definition of the kinetic and potential energy of the system. This family of models can employ both absolute and relative coordinates reduction and they are derived from continuum mechanics theories. When no specific assumptions are made, they are based on the classical three-dimensional continuum mechanics theory, while other approaches have been presented for surface structures (shells, membranes) or slender structures (beam, rods). In particular, the latter include Cosserat, Kirchhoff and non-linear Euler Bernoulli beam theories, which are frequently employed in soft robotics.\\
\noindent 2) \textit{Geometrical models.} They are based on geometrical assumptions on the deformed shape undertaken by the soft body when specific loads are applied. For this group, the central role is taken by the generalized coordinates, on which the system's kinetic and potential energy are defined. In particular, two main groups emerged. The first one includes the ''functional'' models, which all rely on the assumption that the deformed shape of the body resembles a theoretical space curve represented by a specific mathematical function. In this case, the generalized coordinates are usually absolute in nature. The second group includes the widely known piecewise-constant-curvature (PCC) models, which are based on the discretization of the continuous soft body in a finite number of sections having circular arc shape, with intrinsically relative coordinates.\\
\noindent 3) \textit{Discrete material models.} As the name itself suggests, these models are based on a discretization of the continuous body in a finite number of discrete material components. As such, they are characterized by an \textit{a priori} finite-dimensional configuration space of absolute and/or relative coordinates, the relative ones being usually preferred in practice.\\
\noindent 4) \textit{Surrogate models.} The configuration of the system is obtained using sets of data and a learning process. The great majority of the approaches falling in this group use neural networks models and machine learning algorithms.\\

\noindent This classification has been defined to ease the explanation of the models, in an effort to group them based on the different parametrization path that they employ
(Fig. \ref{fig:Reductions}). Clearly, this is only one of the classifications that can be conceived and, in some cases, it is possible to note some overlapping between the models, that we will try to underline while going through them and which are summarized in Sec. \ref{sec::Discussion}.

\section{Continuum Mechanics Models}
\label{sec::Phys}
\noindent Soft robotics is an extremely interdisciplinary field and continuum mechanics represents the most influential community for the theoretical modeling of soft robots. Classical elasticity theories have been used for centuries to precisely model the mechanics of continuum bodies and they offer an established and general framework that is already available to the soft robotics community. Thus, one of the main goals of using these approaches in the robotics community is to make them computationally efficient for the purpose at hand, while maintaining the realism of the models produced.

\noindent In the following, we will start considering in Sec. \ref{sec::3D} the models that are obtained from the three-dimensional elasticity theory, treating the soft body as a continuum medium. 
We will then move in Sec. \ref{sec::Director} to the so-called "director approaches" \cite{Villaggio}. 
For slender bodies, these approaches are all based on the Cosserat rod theory which encompasses the Kirchhoff and the large deflections (non-linear) Euler Bernoulli theories.


\subsection{Classical 3D Models}
\label{sec::3D}
\begin{figure}[h]
\centering
\vspace{-0.3cm}
\includegraphics[width=0.45\columnwidth]{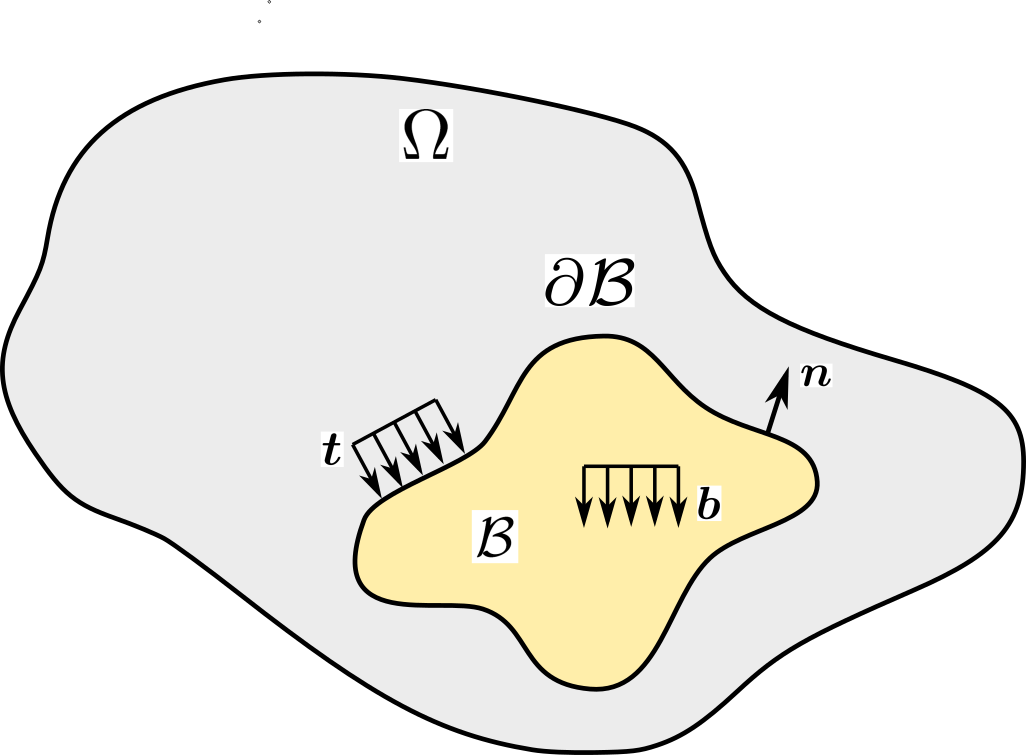}
\caption{\footnotesize Scheme of the considered body}
\label{fig:Cauchy}
\vspace{-0.3cm}
\end{figure}

\noindent Defining a (soft) body as a set of material particles $\Omega$ labelled by their (material, in general curvilinear) coordinates $\bm{X}$, the aim of any three dimensional theory is to predict the time-evolution of its configuration $\bm{r}$ defined by:
\begin{equation}\label{position_3D}
\bm{r}(\cdot): \bm{X} \in \Omega \mapsto \bm{r}(\bm{X}) \in \mathbb{R}^3.
\end{equation}
In order to proceed, the first step consists in obtaining the balance equations of the system. Shall we consider a material subpart of the body $\mathcal{B}\subset\Omega$, having frontier $\partial \mathcal{B}$ and outward unit normal $\bm{n} \in \mathbb{R}^3$, Fig. \ref{fig:Cauchy}. In general, volume and surface forces are applied upon $\mathcal{B}$, and modelled by some volume and surface vector densities, $\bm{b} \in \mathbb{R}^3$ and $\bm{t} \in \mathbb{R}^3$ respectively.
While the first represents the external forces acting on the volume, the latter represents the forces that $\mathcal{B}$ exchanges with its surroundings (the other parts of $\Omega$) across its boundary $\partial \mathcal{B}$, which are also known as internal forces. In particular, $\bm{t}$ is called stress vector and we admit the Cauchy theorem, i.e. $\bm{t} = \bm{\sigma} \bm{n}$, where $\bm{\sigma} \in \mathbb{R}^3 \times \mathbb{R}^3$ is the Cauchy's stress tensor. It is then possible to express the balance equations for any part $\mathcal{B}$ of $\Omega$, as:
\begin{equation}
\label{Balance_sub_domain}
    \int_{\mathcal{B}} \left( \bm{b} - \rho \dot{\bm{v}} \right) dV + \int_{\partial \mathcal{B}} \bm{t} dS = 0,
\end{equation}
where $\rho$ is the mass density, $\bm{v}$ is the body velocity and "$\cdot$" denotes the partial time-derivation $\partial./\partial t$. Using divergence theorem, and remarking that (\ref{Balance_sub_domain}) holds for any material sub-domain $\mathcal{B}$, yields the Cauchy equilibrium equations:
\begin{equation}
\label{Cauchy}
\mathbf{\nabla} \cdot \bm{\sigma} + \bm{b} = \rho \dot{\bm{v}},
\end{equation}
where "$\mathbf{\nabla} \cdot$" stands for the divergence operator. In order to close the formulation, \eqref{Cauchy} needs to be supplemented with a definition of the strains and a constitutive relation characterizing the response of the body's material under external forces (or, in general, any external stimuli), i.e. a relationship between the stress tensor $\bm{\sigma}$ and an objective strain measure as the Green-Lagrange tensor field:
\begin{equation}\label{Green_Lagrange_strains}
\bm{E}=\frac{1}{2}(\mathbf{\nabla}\bm{r}^{T}\mathbf{\nabla}\bm{r}-\mathbf{\nabla}\bm{r}_o^{T}\mathbf{\nabla}\bm{r}_o),
\end{equation}
where $\mathbf{\nabla}\bm{r}$ is the gradient (the Jacobian) of the nonlinear map \eqref{position_3D}, while the index $o$ denotes the positional field over a reference configuration. 
There are different ways to define a constitutive relation and the first and most famous example is the Hooke's law for linear elastic materials, which in the 1D case reads $\sigma = E \epsilon$. One other more general constitutive relation has been proposed by Green, employing the concept of strain energy function. More precisely, a Green elastic material (also known as hyperelastic material) is a type of medium for which the stress–strain relationship can be derived from a strain energy density function (or stored-energy function). This function depends symmetrically on the principal stretches (the eigenvalues of (\ref{Green_Lagrange_strains})) $\lambda_1$, $\lambda_2$ and $\lambda_3$, which, for incompressible materials satisfy the constrain $\lambda_1\lambda_2\lambda_3 =1$. It is worth to remind that the stretch ratios $\lambda_i$ are defined as the ratio between the stretched length $l_i$ and the undeformed one $l_{0,1}$. Many soft robotics devices and components are realized with rubber-like materials, which, in the case of static deformations, are often treated as hyperelastic (while, in general, they can exhibit other nonlinear behaviors such as hysteresis, visco-elasticity and stress softening). Beyond Green's model, the Ogden \cite{Odgen} material model represents one of the most general framework for the modeling of hyperelastic materials. In such a material, the strain energy density is expressed in terms of the principal stretches of the left Cauchy-Green strain tensor $\bm{B}=(\mathbf{\nabla}\bm{r}\mathbf{\nabla}\bm{r}_o^{-1})(\mathbf{\nabla}\bm{r}\mathbf{\nabla}\bm{r}_o^{-1})^{T}$ as:
\begin{equation}
\label{W_Odgen}
    U( \lambda_1, \lambda_2, \lambda_3) = \sum_{j=1}^{n} \frac{\mu_j}{\alpha_j} (\lambda_1^{\alpha_j} + \lambda_2^{\alpha_j} + \lambda_3^{\alpha_j} - 3),
\end{equation}
where $n$, $\mu_j$ and $\alpha_j$ are material constants. In particular, for $n=1$ and $\alpha=2$, the Neo-Hookean model is obtained, while the Mooney-Rivlin model for an incompressible material is obtained for $n=2$, $\alpha_1=2$, $\alpha_2=-2$. Once the strain energy function is defined, the principal Cauchy stresses $\sigma_1$, $\sigma_2$ and $\sigma_3$ are related to the principal stretches through the equations:
\begin{equation}
\label{sigma_Odgen}
    \sigma_i = \lambda_i \frac{\partial U}{\partial \lambda_i} - p,
\end{equation}
where $p$ represents the Lagrange multiplier (or pressure), associated with the incompressibility constraint. It is worth to remind that the Cauchy principal stresses are related to the corresponding nominal (or engineering) stresses $\sigma_e$ by the relationship $\sigma_{e,i} = \sigma_i \lambda_i^{-1}$. The great majority of the materials that are employed within the soft robotics community can be considered as hyperelastic. However, most soft robotics modeling approaches still rely on a linear-elastic material assumption, as they are mostly focused on the description of the large deformation of the body, rather than the large strains.

\subsection{Directors Approaches}
\label{sec::Director}
\begin{figure}[h]
\centering
\includegraphics[width=0.75\columnwidth]{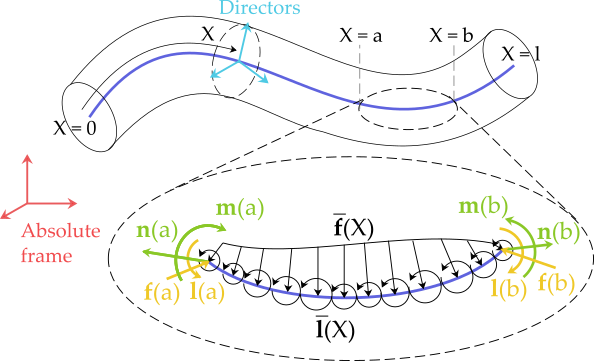}
\caption{\footnotesize Cosserat rod representation}
\label{fig:Cosserat}
\vspace{-0.45cm}
\end{figure}

\noindent The above 3D model applies apriori to material bodies of any geometry (at rest). However, there are many situations where the bodies considered have particular shapes, flattened, or elongated, i.e. where two, or one of the material dimensions, dominate the others. In the first case, we speak of plates or shells, in the second of beams or arches. In these cases, the full 3D position field of (\ref{position_3D}) can be developed in Taylor series around a reference material surface for shells, or a material line for beams, and the dependence on small orthogonal material dimensions analytically pre-integrated into the thickness. At leading order, this provides reduced models where the positional field $\bm{r}$ of (\ref{position_3D}) is replaced by that of the reference surface or line only, plus vector fields supporting the remaining small dimensions and named \emph{directors}. For both 2D or 1D reduced media, the resulting models are due to Kirchhoff and Reissner depending whether transverse shearing is neglected or not. In robotics, except for modelling soft robots inspired of octopus \cite{Renda_BB2015}, \cite{Boyer_2017_Shell}, \cite{Renda_IJRR2018}, the great majority of these reduced models are beam models. To obtain such models, one can advantageously replace the limit process applied to the 3D model \cite{Villaggio}, by a simpler approach consisting in directly modeling the rod as a material line along which is continuously stacked a set of rigid cross-sections (supporting the directors), and labeled by a single material coordinate $X$. In this case, the rod is called a \textit{Cosserat rod} (Cosserat and Cosserat, 1907 \cite{Cosserat}), and its configuration is defined by the positional vector field $\bm{r} \in \mathbb{R}^3$ of its center-line, and the orientation (rotation) field $\bm{R} \in SO(3)$, of its (mobile) cross-sectional basis (of directors). Both these fields are functions of $X \in [0,l]$, where $l$ is the rod length at rest. With this parametrization, the balance equations of the rod can be derived with Newton's law and Euler's theorem, by considering an arbitrary subinterval $[a,b] \subset [0,l]$ of the rod, Fig. \ref{fig:Cosserat}.
Across any $X$-section splitting the rod in two parts, the action of the side $Y \leq X$ upon the other side is modeled by the stress resultant vector $\bm{n}(X)$ and the couple stress $\bm{m}(X)$ in the global frame. The mass density of the rod is $\rho$, $\bm{I}$ is its cross-sectional inertia tensor, while $\bm{v}$ and $\bm{\omega}$ are the linear and angular velocities of cross-sections in the global frame. The rod is also subject to some external forces $\bm{\bar{f}}$, and moments $\bm{\bar{l}}$ densities per unit of $X$, in the global frame. Stating the linear and angular (Newton-Euler) balance equations over the subdomain $[a,b]$, and 
taking the limit of the interval going to $0$, 
provides the dynamic equilibrium \cite{Villaggio}:
\begin{equation}
\label{CR}
\left\{
\begin{array}{l}
\bm{n}' + \bm{\bar{f}} = \rho A \dot{\bm{v}}, \vspace{0.05cm}\\
\bm{m}' + \bm{r}' \times \bm{n} + \bm{\bar{l}} = \partial_{t}(\rho \bm{I} \cdot \bm{\omega}), \vspace{0.05cm} \\
\bm{n}(0) = -\bm{f}_0 \; \text{,} \hspace{0.15cm} \bm{n}(l) = \bm{f}_{l} \; \text{,}
\hspace{0.1cm} \bm{m}(0) = -\bm{l}_0 \; \text{,} \hspace{0.15cm} \bm{m}(l) = \bm{l}_{l} \; \text{,}
\end{array}
\right.
\end{equation}
where from now on, a "$'$" denotes the space-partial derivative $\partial./\partial X$, while $\bm{l}_{0,l}$ $\bm{f}_{0,l} $ are the moments and forces applied in $X=0$ and $X=l$, respectively. In order to close the formulation, \eqref{CR} needs to be supplemented with a definition of strains and a stress-strain law. In the assumption of finite deformations and small strains, one can use as strain measurements the two $3\times 1$ vectors:
\begin{equation}\label{Cosserat strains}
\bm{\epsilon}_{\ell}=\bm{R}^{T}(\bm{r}'-\bm{r}_o')\,\,,\,\, \bm{\epsilon}_a=\bm{K}-\bm{K}_o,
\end{equation}
where the index $o$ denotes a reference configuration of the rod, $\bm{K}=(\bm{R}^{T}\bm{R}')^{\vee}$ is the vector of material torsion and curvature in the cross-sectional frame, where $\bm{\vee}$ changes any skew symmetric matrix $\bm{W}\in \mathbb{R}^{3}\times \mathbb{R}^{3}$ into $\bm{W}^{\vee}\in \mathbb{R}^{3}$ s.t. $\forall\,\bm{V}\in \mathbb{R}^{3}$, $\bm{W}\bm{V}=\bm{W}^{\vee}\times \bm{V}$, while conversely $\wedge$ is s.t. $\bm{\hat{W}}^{\vee}=\bm{W}$. In this context the constitutive law takes the usual linear Hookean form:
\begin{equation}\label{KIRCH}
\bm{n} =\bm{R}\bm{\mathcal{H}}_{\ell} \bm{\epsilon}_{\ell}\,\,,\,\,\bm{m} =\bm{R} \bm{\mathcal{H}}_a \bm{\epsilon}_a,
\end{equation}
with $\bm{\mathcal{H}}_{\ell}=\bm{diag}(EA, GA, GA)$ and $\bm{\mathcal{H}}_a=\bm{diag}(GI_1, EI_2,$ $ EI_3)$ defining the usual cross-sectional linear and angular stiffness matrices in the cross-sectional frames respectively. In the recent years, the above model has been reformulated and exploited in the modern language of Cosserat rod theory on Lie groups. Introduced in bio-robotics \cite{Boyer_TRO_2006}, \cite{Boyer2012} to study  dynamics of hyper-redundant locomotors, in this setting, the configuration of a soft body is directly defined as a curve:
\begin{equation}
\label{conf_Cosserat}
\bm{g}(\cdot): X \in [0, l] \mapsto \bm{g}(X) \in SE(3),
\end{equation}
where $\bm{g}(X)$ denotes the $4\times 4$ usual homogeneous matrix of the $X$-cross sectional frame, with positional and rotational components denoted $\bm{r}(X)$ and $\bm{R}(X)$ respectively. The space variations of the field $\bm{g}$ can be entirely described by the field of space-twists $\bm{\xi}=\left(\bm{K}^T, \bm{\Gamma}^T \right)^T\in \mathbb{R}^6$, which defines a continuous geometric model of the rod:
\begin{equation}
\label{gprimo}
    \bm{g}' = \bm{g} \widehat{\bm{\xi}} \; , \hspace{0.3cm} \bm{g}(0) = \bm{g}_0.
\end{equation}
With these definitions, the strain state of (\ref{KIRCH}) can be re-expressed as a unique field of twist $\bm{\epsilon}=(\bm{\epsilon}_a^{T},\bm{\epsilon}_{\ell}^{T})^{T}=\bm{\xi}-\bm{\xi}_{o}$. Similarly, replacing $X$ by the time variable $t$, yields:
\begin{equation}
\label{diff_kin}
\dot{\bm{g}} = \bm{g} \bm{\widehat{\eta}} \; \text{,}
\end{equation}
which defines the field of velocity twist $\bm{\eta} = \left(\bm{\Omega}^T, \bm{V}^T \right)^T \in \mathbb{R}^6$, where $\bm{\Omega}(X), \bm{V}(X) \in \mathbb{R}^{3}$ are the angular and linear velocity of the $X$-cross-sectional frame in its mobile basis, which are related to their inertial counter-part through: $\bm{\Omega} = \bm{R}^T\bm{\omega}$, $\bm{V} = \bm{R}^T\bm{v}$. 
On $SE(3)$, the dynamic equilibrium equations of a rod subject to a density of external wrench $\bar{\bm{F}}$ on $ ]0, l[$ and two tip external wrenches $\bm{F}_0$ and $\bm{F}_{l}$ at $X=0$ and $X=l$ respectively, are expressed in the cross-sectional frames, in the form \cite{Boyer2012},  \cite{Renda_TRO2014}:
\begin{equation} \label{SE(3)_Cosserat_equilibrium}
\left\{
\begin{array}{l}
 \bm{\mathcal{M}} \dot{\bm{\eta}} - \mathrm{ad}_{\bm{\eta}}^T  \bm{\mathcal{M}} \bm{\eta} = \bm{\Lambda}' - \mathrm{ad}^{T}_{\bm{\xi}} \bm{\Lambda} + \bar{\bm{F}}  \; \text{,}  \\
 \bm{\Lambda}(0) = -\bm{F}_0  \; \text{,} \hspace{0.2cm} \bm{\Lambda}(l) = \bm{F}_{l},
\end{array}
\right.
\end{equation}
where $\mathrm{ad}$ is the adjoint action of the Lie algebra, $ \bm{\mathcal{M}}$ is the $6\times6$ cross sectional inertia matrix and $\bm{\Lambda}$ models the stress field along the beam (it is the dual counterpart of the strain field). 
The balance \eqref{SE(3)_Cosserat_equilibrium} can be derived by applying Hamilton's principle in the framework of Lagrangian reduction theory \cite{Boyer_2017_Shell}, or deduced from \eqref{CR}, by using the relations between vectors of components in the inertial and cross-sectional frames: $\bm{\eta}  = ((\bm{R}^T\bm{\omega})^T, (\bm{R}^T\bm{v})^T)^T$,  $\bm{\Lambda} = ((\bm{R}^T\bm{m})^T, (\bm{R}^T\bm{n})^T)^T$, $\bm{F}_{0,l}=((\bm{R}^T \bm{l})_{0,l}^T, (\bm{R}^T \bm{f})_{0,l}^T)^{T}$, and $\bar{\bm{F}} = ((\bm{R}^T \bm{\bar{l}})^T, (\bm{R}^T \bm{\bar{f}})^T)^{T}$. Equations \eqref{SE(3)_Cosserat_equilibrium} are first order PDEs that govern the time-evolution of velocities $\bm{\eta}$ along the rod. Thus, they need to be supplemented with the kinematic model (\ref{diff_kin}) allowing to reconstruct (by integration) the time-evolution of the configuration $\bm{g}$.

\noindent To apply the Cosserat rod model to soft robots, an actuation model must be introduced into the above formulation. In general, this has to be done on a case by case basis, depending on the specific technology and design of the robot. However, this task has been fully accomplished in the case of tendon-driven robots, whose design is now stabilized. Two equivalent approaches have been proposed in this case, depending whether the effect of actuation is introduced through the densities of external wrenches $(\bar{\bm{f}},\bar{\bm{l}})$ or through a field of internal stress $\bm{\Lambda}$ added to the elastic restoring ones of the constitutive law \eqref{KIRCH}. In both approaches the tendons are idealized as inelastic and friction-less force transmitters. Initiated in the planar case in \cite{Camarillo_TRO2008} and extended to the three-dimensional case in \cite{Rucker2011}, the first approach is based on the application of Newton's laws to the rod and to each of the tendons, isolated separately. Then, using the action-reaction principle provides the internal forces and moments exerted by a set of $n_a$ tendons onto the rod:
\begin{equation}
\label{tendon_actuation_as_external_wrenches}
\left(
      \begin{array}{c}
        \bar{\bm{l}} \\
        \bar{\bm{f}} \\
      \end{array}
    \right)= - \sum_{i=1}^{n_a} \left(
      \begin{array}{c}
        \bm{d}_i\times\bm{t}_{ci}' \\
        \bm{t}_{ci}' \\
      \end{array}
    \right)u_i,
\end{equation}
where $\bm{d}_i(X)$ and $\bm{t}_{ci}(X)$ are, respectively, the position vector of the tendon routing $i$ with respect to the rod's backbone and its unit tangent vector, both expressed at $X$, while $u_i(t)$ is the uniform tension ($u_i'=0$) transmitted along the tendon. In the second approach \cite{Renda2017}, \cite{Boyer_TRO_2020}, applying the virtual work principle to the whole arm, allows modelling the effect of tendons by changing the constitutive law \eqref{KIRCH} into the active one \cite{Boyer_TRO_2020}:
\begin{equation}
\label{active_constitutive_law}
    \left(
      \begin{array}{c}
        \bm{m} \\
        \bm{n} \\
      \end{array}
    \right)
     = \left(
         \begin{array}{c}
           \bm{R}\bm{\mathcal{H}_a}\bm{\epsilon}_a \\
           \bm{R}\bm{\mathcal{H}_{\ell}}\bm{\epsilon}_{\ell} \\
         \end{array}
       \right)
      + \sum_{i=1}^{n_a}
        \left(
        \begin{array}{c}
         \bm{d}_i \times \bm{t}_{ci} \\
        \bm{t}_{ci} \\
        \end{array} \right)
    u_i,
\end{equation}
where the derivatives of $\bm{t}_{ci}(X)$  appearing in \eqref{tendon_actuation_as_external_wrenches} are now taken in charge by the space derivations of the equilibrium equations \eqref{SE(3)_Cosserat_equilibrium}.

\noindent Remarkably, other nonlinear beam theories can be defined as sub-models of the above theory, often referred to as the Reissner beam model \cite{Reissner_1972}. The Kirchhoff rod theory can be obtained by fixing in $\bm{\xi}$ of (\ref{gprimo}), the first component $\Gamma_1$ of $\bm{\Gamma}$ to one, and the two others to zero, which prevents the beam from stretching and shearing respectively. By imposing that the motion of a Kirchhoff rod is planar and ignoring its thickness, the stress equilibrium (\ref{SE(3)_Cosserat_equilibrium}), restricted to statics, can then be integrated explicitly. The model is then that of the \textit{elastica}, i.e. an elastic line of minimal curvature energy, and we have:
\begin{equation}
\label{EB}
    EI(X)\theta' = M_{ext},
\end{equation}
where $\theta'(X)$ is its curvature in the plane, $EI(X)$ its bending stiffness, and $ M_{ext}(X)$ is the integral of the external bending moments about the center of the $X$-cross section, acting on the remaining part ($Y>X$) of the beam.

\section{Numerical resolution of 3D continuum formulations}
\label{sec::Num3D}
\noindent To integrate the dynamics of a soft manipulator modelled by 3D continuum mechanics, one needs to solve for each material sub-domain a closed formulation obtained by the PDEs (\ref{Cauchy}) with a definition of strains, a constitutive law of the type of (\ref{sigma_Odgen}) and BCs (some of them imposed by the connection of the bodies). The usual numerical resolution methods use finite differences to discretize the time axis through explicit or implicit integration schemes. However, the spatial discretization can be achieved differently by using finite differences, boundary elements or finite elements method. The first approach is based on Taylor approximations of the PDEs, but it is difficult to apply to complex shaped objects, while the second is based on Green's theorems and restricted to linear problems. In contrast, the finite elements method has the advantage of inheriting the variational structure of the original formulation, which is transformed into a robust nonlinear optimization problem in the context of the Ritz method.

\subsection{Finite Element Method (FEM) for 3D formulations}
\label{sec::FEM}
\noindent FEM is a well-known and widely-spread numerical technique for finding approximate solutions to Partial Differential Equations (PDEs). It was established by a set of scientific papers in the 40s and it soon became one of the most commonly employed technique for the modeling of a wide range of engineering problems, from structural mechanics to fluid flow and heat transfer. The main property of this technique is the sub-division of the problem's domain into a set of smaller parts, called \textit{finite elements}. These are obtained by the construction of a \textit{mesh} (Fig. \ref{fig:Reductions}(a)) which represents the numerical domain for the solutions. Once this space discretization is performed, the continuous field $\bm{r}$ is approximated on each element by a polynomial interpolation of its values at the edges of the elements, called nodes, according to a Ritz reduction approach. Introducing this discretization in the above continuous formulation and projecting it on the same trial (polynomials) basis, changes the PDEs \eqref{Cauchy} into a set of algebraic discrete equations for steady state problems, or ODEs for transient problems. In the former case, the static equilibrium becomes:
\begin{equation}
\label{FEM_static}
    \bm{Q}_{int}(\bm{q}) = \bm{Q}_{ext}(\bm{q}),
\end{equation}
where $\bm{Q}_{int}(\bm{q})$ and $\bm{Q}_{ext}(\bm{q})$ represent the internal and external generalized forces, respectively, while $\bm{q}$ are the generalized coordinates, i.e. the nodal positions. On the other hand, the dynamic equilibrium is given by a system of ODEs:
\begin{equation}
\label{FEM_dynamic}
\left\{
\begin{array}{l}
   \dot{\bm{q}} = \bm{v},
   \vspace{0.05cm}\\
    \bm{M}\dot{\bm{v}} + \bm{Q}_{int}(\bm{q}, \bm{v})  = \bm{Q}_{ext}(\bm{q}, \bm{v}),
\end{array}
\right.
\end{equation}
where $\bm{M}$ is the (constant) generalized mass matrix. Note that the terms appearing in \eqref{FEM_static} and \eqref{FEM_dynamic} are computed integrating the distributed variables element-by-element according to the selected shape functions, which play the role of the kinematic map, while the position of each element in the mesh is ensured by the assembly process. Since nodal coordinates are of absolute nature, the non-linearities appear in $\bm{Q}_{int}(\bm{q}, \bm{v})$. They are named \textit{geometric} or \textit{material}, depending whether they come from the large displacements (and namely the rotations) of the elements, or from the constitutive laws. To capture the local finite rotations along bodies, the corotational approach is often used in three-dimensional FE software. One other solution to represent the non-linearities is to use an Absolute Nodal Coordinate Formulation (ANCF-FEM) \cite{Shabana_SORO2018} .

\noindent The FEM represents a standard for many popular simulation software, such as Abaqus, ANSYS, Comsol. The main advantage of these software is that they provide a powerful and general tool that can be applied to a wide spectrum of physical problems ranging from structural dynamics, fluid-structure interactions, contact and thermodynamics. Moreover, they provide a ready made and user-friendly framework that can be easily employed  for specific studies on multiphysics systems. They have been widely used to model soft robots \cite{Xavier2021}, \cite{Moseley2016} (Abaqus), \cite{Tawk2020} (ANSYS). At the same time, their generality intrinsically entail an increased computational cost and the simulations, especially in dynamics, can take a lot of time to converge.
There are some popular examples of ad-hoc Finite Element approaches (and softwares) that have been specifically proposed for (soft) robotics applications. In 2007 \cite{Allard_2007}, a group of researchers from different institutes released Sofa (Simulation Open Framework Architecture), an open-source C++ library which was originally presented as a computational environment for medical simulations \cite{Sofa2012}, \cite{Coevoet_2017}. In the following years, Sofa became a comprehensive high-performance library that have been widely implemented for different application fields, and a SoftRobots plugin was also created for the design \cite{Zheng_2019}, modeling and control \cite{Duriez_ICRA2013}, \cite{Duriez_TRO2018} of soft robots, including self-collision scenarios \cite{Goury_RAL2021}. More recently, a model combining the FEM and the discrete Cosserat approach of Sec. \ref{sec::Director} was presented in \cite{Adagolodjo_RAL2021}.
A discussion on the architecture of the software is addressed in Sec. \ref{sec::Comput}, while here we mostly focus on the modeling part for the soft bodies. In Sofa, a deformable continuum is modeled using a dynamic or quasi-static system of simulation nodes. The node coordinates are the independent DOFs of the object, and they are typically governed by equations of the type \eqref{FEM_static} and \eqref{FEM_dynamic}. Some approaches have also introduced reduced coordinates and these are presented in Sec. \ref{sec::LOR}. With regards to the actuation, a constraint-based approach is employed. Imposing a linearization of the internal forces $\bm{Q}_{int}(\bm{q}_i) \approx \bm{Q}_{int}(\bm{q}_{i-1}) + \bm{K}(\bm{q}_{i-1})d\bm{q}$ and considering the contribution of the actuation constraints, the static equilibrium at each $i$-th step \eqref{FEM_static} becomes:
\begin{equation}
\label{eq_Sofa}
    \bm{Q}_{int}(\bm{q}_{i-1}) + \bm{K}(\bm{q}_{i-1})d\bm{q} = \bm{Q}_{ext}(\bm{q}_{i-1}) + \bm{B}(\bm{q}_{i-1})^T\bm{\lambda},
\end{equation}
where the term $\bm{B}^T \bm{\lambda}$ models the contributions of the Lagrange multipliers (i.e. those of the actuators), and $\bm{B}$ represents the Jacobian of the constraint equations imposed by the actuators.
Three steps are then performed: (1) a free configuration $\bm{q}^{free}$ is found for $\bm{\lambda}=0$ and, for each constraint, the violation $\bm{\delta}^{free}$ is estimated; (2) the solver computes the value of $\bm{\lambda}$ through a projection of the mechanics into the constraint space, obtaining the smallest possible projection space $\bm{\delta} = \bm{B}\bm{K}^{-1}\bm{B}^T\bm{\lambda} + \bm{\delta}^{free}$; (3) the final configuration is corrected using the value of the constraint response $\bm{q} = \bm{q}^{free} + \bm{K}^{-1}\bm{B}^T\bm{\lambda}$.
For a fluidic actuation, $\bm{\lambda}$ and $\bm{\delta}$ represent the cavity pressure and volume. For a cable, $\bm{\lambda}$ and $\bm{\delta}$ are the cable tension and length. While in the inverse model case, $\bm{\lambda}$ is unknown and can be find by solving a QP type problem \cite{Coevoet_2017}, in the direct problem case, $\bm{\lambda}$ can be unknown too, as this is the case of a cable driven arm where the non-negative tensions condition is modelled as a set of unilateral and complementarity (non-holonomic) constraints. 
Because standard finite elements are not differentiable at element boundaries, some authors have recently implemented a Moving Least Squares (MLS-FEM) formulation, making the deformation gradient twice differentiable \cite{Moritz_2021}, with the aim of applying usual optimization gradient descent techniques to different design problems related to automated actuation routing \cite{Maloisel_TRO2021} and sensor design \cite{Tapia_SoRo2020} for soft robots.

\section{Numerical resolution of directors-based formulations}
\label{sec::Num1D}
\noindent The closed formulation of a soft manipulator modelled by Cosserat rods consists, for each rod, of the kinematic model \eqref{diff_kin}, the stress balance \eqref{SE(3)_Cosserat_equilibrium} (where the BCs can depend on the connection between the rods), and a constitutive law of the type of \eqref{KIRCH}, with the associated strain definition (\ref{Cosserat strains}). This formulation is consistent with the definition of the configurations \eqref{conf_Cosserat} in terms of the absolute fields $\bm{g}$. For soft manipulators, one can alternatively parameterize the configuration with the fields $\bm{\xi}$ or $\bm{\epsilon}$ and use \eqref{gprimo} as a space-reconstruction equation of $\bm{g}$ along the rod. Time-differentiating twice this geometric model \eqref{gprimo} provides two further models \cite{Boyer2012} \cite{Renda_TRO2018}:
\begin{equation}
\label{transp}
\begin{array}{l}
\bm{\eta}^{\prime}  = \dot{\bm{\xi}} - \mathrm{ad}_{\bm{\xi}}\bm{\eta} \; \text{,} \vspace{0.05cm}
\\
\dot{\bm{\eta}}^{\prime} = \ddot{\bm{\xi}} - \mathrm{ad}_{\dot{\bm{\xi}}}\bm{\eta} - \mathrm{ad}_{\bm{\xi}}\dot{\bm{\eta}} \; \text{,}
\end{array}
\end{equation}
whose spatial integration allows to reconstruct the absolute velocity and acceleration fields along the rod, from the strain velocities and accelerations $\dot{\bm{\xi}}$, and $\ddot{\bm{\xi}}$. In this case, the parametrization of configurations is said to be relative and opposed to the classical absolute one based on $\bm{g}$. In this section, we discuss the most important examples of approaches using Cosserat rod's theory for modeling soft robots. They are separated into "non-energetic" and "energetic" methods. In the first case, the original formulation is first set in the form of a space Boundary Value Problem (BVP) with BCs partly defined at one end, and partly at the other. Then, this BVP is directly discretized on a spatial grid with no further reduction. In the second case, often referred to as the "Ritz" or "Ritz-Galerkin" method, the configuration needs first to be parameterized by some vector fields (absolute or relative) that are reduced on a truncated functional basis of space-dependent vectors. The components along these vectors define a finite set of generalized coordinates governed by a set of Lagrange ODEs in time. In the following, Ritz methods are classified into nodal and modal, depending whether the basis is defined by some nodal interpolation polynomials over the finite elements of a mesh, or defined over the full rod domain.

\subsection{Non-energetic approaches}
\noindent In practice, these approaches consist in extracting from the previous model a closed formulation (absolute, relative or mixed) which is reshaped into one of the standard forms of numerical analysis.

\subsubsection{Finite Differences}
\label{sec::FiniteDifferences}
\noindent The finite difference method is one of the simplest and oldest methods to solve partial differential equations. In the case of Cosserat beams, it has been widely used by the ocean engineering community to predict the temporal evolution of towed submarine cables, subjected to low tensions that cause a dynamic singularity in the standard catheter model. Developed by Burgess and his successors \cite{Burgess1993}, \cite{Tjavaras1998}, the direct dynamics of the cable modeled by a shear-less Cosserat rod is formulated as a nonlinear BVP in space and an IVP in time. The application of a spatial finite difference scheme transforms this continuous formulation into a set of time-ODEs at the grid nodes, which can then be solved explicitly, or by a predictor-corrector strategy (with correction by the Newton-Raphson algorithm), depending on whether an explicit or implicit time finite difference scheme is used.
In continuum and soft robotics, finite differences method has been applied to both statics \cite{Peyron_2018}, 
\cite{Briot_2021} and dynamics \cite{Trivedi2008}, \cite{Renda_TRO2014}. One canonical example of such a numerical resolution applied to soft robotics is presented in \cite{Renda_TRO2014}, where the dynamic modelling and simulation of a cable driven multibending soft robot arm is addressed. Gathering the geometric model \eqref{diff_kin}, with that of velocities \eqref{transp}$_1$, and the balance of stress \eqref{SE(3)_Cosserat_equilibrium}, in which the model of tendons actuation \eqref{tendon_actuation_as_external_wrenches} and the constitutive law \eqref{KIRCH} are introduced, allows first to set the forward dynamics BVP in the form $\dot{\bm{z}}=\bm{f}(\bm{z},\bm{z}'\bm{z}'',t)$, where $\bm{z}$ is composed of the angular and linear components of fields $\bm{g}$, $\bm{\epsilon}$ and $\bm{\eta}$, while the time dependency is due to the actuation. Then, approximating, the space derivatives of the kinematic field $\bm{\eta}$ with a forward (from the base to the tip) finite difference scheme, and those of the strains field $\bm{\epsilon}$, with a backward one (from tip to the base), this BVP is changed into a finite-dimensional state system of the usual form $\dot{\bm{x}}=\bm{f}(\bm{x},t)$, that can be time-integrated with a standard explicit Runge-Kutta integrator.

\subsubsection{Shooting Methods}
\label{sec::Shooting}
\noindent The shooting method represents one of the most popular numerical approach to solve a BVP along one dimension of space or time. In the context of Cosserat rods, it has been introduced in the Oceanic Engineering community of submarine cables as an alternative way to finite differences for solving the forward dynamics of these systems \cite{SUN1996}. In the case of soft arms, the archetypal BVP is that of the forward static model of a Cosserat rod. It is obtained by gathering \eqref{gprimo} with the stress balance \eqref{SE(3)_Cosserat_equilibrium}, in which we remove the velocities and accelerations. Then, expressing the strains in terms of stress by inverting the constitutive law \eqref{KIRCH} provides a (forward static) BVP in the form $\bm{z}'=\bm{f}(\bm{z})$, that can be solved by the shooting algorithm, i.e. by integrating a sequence of initial (proximal) value problems, whose unknown proximal conditions are first predicted and then iteratively corrected (with a Newton type-algorithm), until a solution of the BVP that fulfills the known distal boundary conditions is found, Fig. \ref{fig:Reductions}(b).

\noindent In \cite{Rucker2011}, Cosserat theory is used for static and dynamic modeling of continuous robots, driven by tendons. The effect of tendons on the rod is modelled as some densities of external forces and moments defined by (\ref{tendon_actuation_as_external_wrenches}), in which the routing path of each tendon is expressed as a function of the rod strain variables. After some clever algebraic manipulations, an explicit BVP of the forward statics in the form $\bm{z}'=\bm{f}(\bm{z},t)$ is obtained, to which the shooting method is applied for the purpose of calculating the static equilibrium configurations of these systems.
In \cite{Rucker2019} the shooting-based resolution was extended from statics to the forward dynamics of several designs of continuum robots for simulation purposes. Inspired by \cite{SUN1996} and \cite{FlyFishing2002}, an implicit time-integrator is used in order to remove the time derivatives of the system of PDEs \eqref{diff_kin}, \eqref{transp}$_1$, and \eqref{SE(3)_Cosserat_equilibrium}. Using the constitutive law \eqref{KIRCH} and the model of actuation \eqref{tendon_actuation_as_external_wrenches}, provides a BVP in space in the usual form $\bm{z}'=\bm{f}(\bm{z},t)$, that is solved at each time-step of the simulation by the shooting method starting from a prediction. With the rapid expansion of continuum robotics, the approach has been successfully extended to several multi-segment designs, ranging from multi-segment tendon robots \cite{Sharifi2021}, to parallel continuum robots \cite{Black_TRO2018}, \cite{Rucker2019}, to concentric tube robots \cite{Rucker2010} and \cite{Till_TRO202}. In the first case, the BVPs along segments are connected in series by their boundary conditions, and the approach is reapplied in cascade from the base to the tip of the arm. In the second case, it is applied along each legs and the residual vector of distal conditions must integrate the equilibrium equations of the rigid platform. In the third case, the superimposed tubes must be considered as non-material domains of varying length, along which the method is applied in cascade.

\subsubsection{Continuous Newton-Euler based Method}
\label{sec::Newton-Euler}

\noindent The Newton-Euler formalism of rigid multibody systems has produced fast and simple algorithms for solving the direct and inverse dynamics of rigid robots \cite{Featherstone}. In \cite{Boyer_TRO_2006}, the Luh "computed torque algorithm" has been extended to solve the inverse dynamics of fish-inspired hyperredundant robots modeled as Cosserat rods whose shape is controlled by the internal torque field $\bm{R}^T\bm{m}$. Concatenating \eqref{SE(3)_Cosserat_equilibrium} with \eqref{gprimo}, and its first and second time-derivatives \eqref{transp}, defines a continuous Newton-Euler model, where the rod sections take the place of the rigid bodies and $\bm{K}$ that of the joint variables of a rigid discrete system. Imposing the curvature time law in this model provides an inverse dynamics BVP, which, in contrast to the forward dynamics one, can be solved in two decoupled pass (i.e. without resorting to shooting). Solving the (locomotion) dynamics of net motions, this algorithm has been extensively exploited for the study of bio-inspired swimming \cite{Boyer_TRO_2008}. Although first designed for hyperredundant systems, this algorithm has been recently applied to soft and continuum robotics \cite{Boyer_TRO_2020}. Remarkably, when pose-dependent external forces (such as gravity) are neglected, the forward static BVP enjoys the same decoupling property as the inverse BVP, a feature that was exploited in \cite{Renda_BB2012} for quasi-static simulation of tendon-driven robots with the same two pass approach. Note that in such cases, the Newton-Euler and finite difference based approaches can lead to the same numerical solutions but from different points of view.

\subsubsection{Collocation Methods}
\label{sec::Collocation}
\noindent Alternatively to shooting, the forward static BVP of soft arms can also be solved using the collocation method. The main idea is to replace the unknown strain field with a polynomial and setting the vanishing of the residual between the two in a finite set of points on the domain, also called collocation points. Considering $m$ collocation points, a total of $m+1$ equations is obtained, providing the parameters that are required to define an $m$-th order polynomial. In \cite{Simaan2020}, the ODEs describing the kinematics of a Cosserat rod are directly solved in terms of transformations on $SE(3)$, using a combination of orthogonal collocation and forward integration through Magnus Expansion on $SE(3)$. In particular the unknown $\bm{\xi}(X)$ in \eqref{gprimo} is expressed as a set of three Chebyshev polynomials.

\subsection{Energetic approaches}
\label{sec::EnergeticA}
\noindent Energetic approaches seek to reduce the continuous models of Section III.B to a finite set of Lagrangian ODEs in time, thus retaining the variational (energetic) structure of the modeling.

\subsubsection{Absolute nodal-Ritz reduction: GE-FEM}
\label{sec::GEFEM}
\noindent One of the most powerful method to capture in an exact manner the geometric nonlinearities of soft robots is the geometrically-exact FEM (GE-FEM) introduced by J.C. Simo  \cite{SIMO_1988}. In this approach, the model of rotations is introduced at the same level as that of positions, by replacing the material points of classical medium by the rigid micro-structures of a Cosserat medium. The GE character, then imposes to apply the FEM without resorting to any simplifications on finite rotations, except the unavoidable space and time discretizations required by the numerical resolution. The GE-FEM was developed for both shells \cite{SIMO_1989}, and rods \cite{Cardona_1988}, \cite{Boyer_2004}, \cite{Sonneville_2014}. The method’s tour de force was to generalize all the key operations of the FEM (interpolation and space and time-integration), from the linear vector space of positional fields to the curved manifold $SO(3)$. Rooted in Lagrangian mechanics, the resulting dynamic equations take a form similar to that of \eqref{FEM_dynamic}, except that $\bm{q}$ and $\bm{v}$ are now defined by the sets of $(\bm{r},\bm{R})$ and $(\dot{\bm{r}},\bm{\Omega})$ at the nodes, while an additional Coriolis term survives due to the intrinsic curvature of $SO(3)$:
\begin{equation}
\label{general_dynamic}
\bm{M}\dot{\bm{v}} + \bm{C}(\bm{v})\bm{v} + \bm{Q}_{int}(\bm{q}, \bm{v})  = \bm{Q}_{ext}(\bm{q}, \bm{v}).
\end{equation}
Finally, \eqref{general_dynamic} is solved iteratively at each time step according to a standard prediction-correction strategy, in which the time-dependence is removed with an implicit geometric time integrator on $SO(3)\times \mathbb{R}^3$, which preserves the Lie group structure and the geometrically-exact character of the approach. This approach, which is probably the most advanced for modeling mechanisms of rods subject to rigid motions and finite deformations, is today used as a reference for others, and commercialized in the SAMCEF software through its plugin MECANO \cite{MECANO}. In this software, the model of rods is the full-Cosserat model of Reissner. However, its dynamic resolution is ill-conditioned when the aspect ratio of the rod increases, i.e. for very slender rods. To overcome this limitation, several FEM based on the non-linear Euler-Bernoulli or Kirchhoff rod models, were proposed, such as the Absolute Nodal FEM \cite{Shabana_1997} or the Kirchhoff GE-FEM \cite{Boyer_2004}, \cite{Meier2019}. Despite their power, these methods have not yet been adapted to the specific needs of soft robotics as was done for Sofa. As a result, specific applications, more or less related to the original spirit of Simo's GE-FEM were proposed for these systems. In \cite{Tunay2013}, a GE-FEM approach for modeling inflatable robots is presented. A strain field measuring radial inflation is added to the usual 6 strains of the Reissner model. Finally, the weak form of the static equilibrium in this augmented space is derived and solved with a FEM software (COMSOL).

\subsubsection{Absolute modal-Ritz reduction}
\label{sec::AmodalRitz}
\noindent Ritz method can be applied directly on the full domain of the rod. In this case, we refer it to the modal Ritz method. The approach can be applied both to absolute or relative vector fields. In the first case, one needs first to parameterize the rod configuration with a set of vector fields (e.g. the positional field along the backbone $\bm{r}$ and the vector field of its 3 cross-sections Euler's angles $\bm{\theta}$). Then, applying the separation of variables, the components of these fields can be approximated on a truncated basis of Ritz functions compatible with the boundary conditions:
\begin{equation}\label{RitzGal}
   \bm{r}(X,t) = \bm{\Phi}(X) \bm{q}_r(t) \,,\,\bm{\theta}(X,t) = \bm{\Psi}(X) \bm{q}_\theta(t),
\end{equation}
where $\bm{\Phi}(X)$ and $\bm{\Psi}(X)$ are matrices of spatial shape functions, or "modes" (stacked in columns), while $\bm{q}=(\bm{q}_r^{T},\bm{q}_\theta^{T})^{T}$ is a vector of time-dependent generalized (modal) coordinates.

\noindent In \cite{Sadati_RAL2018}, the Ritz-Galerkin method is used to reduce the weak-form of a Cosserat rod in both statics and dynamics. The kinematics of the rod is obtained expressing equation \eqref{gprimo} in an inertial reference system, placed at the base of the manipulator. Then, the positional field of the backbone can be approximated as in \eqref{RitzGal}, while using Kirchhoff assumption and Bishop frames, the rotation field is parameterized with a unique further field of torsion angle. Applying \eqref{RitzGal} to this field of angle and introducing all these approximations into the static balance equations projected onto the same truncated basis, provides a set of algebraic equations in the usual form \eqref{FEM_static}. Using a numerical root finder finally provides the $q_i$s that minimize the approximation error. In the dynamic case, applying the same reduction process provides ODEs of the form \eqref{general_dynamic} but where $\bm{M}$ and $\bm{C}$ now depends on $\bm{q}$ due to the geometric nonlinearities introduced by the parametrization of $SO(3)$. The coefficients $q_i(t)$ are then found by explicit time-integration of these ODEs.

\subsubsection{Relative nodal-Ritz reduction}
\label{sec::RnodalRitz}
\noindent After the works of Simo, a small group of authors proposed an alternative GE-FEM based on the polynomial interpolation of strains instead of poses \cite{Cesarek_FEAD2013}. 
This approach is more accurate, but also more computationally complex than conventional GE-FEM. In the field of soft robotics, a GE-FEM based on strain parametrization was presented in \cite{Grazioso_SORO2019}. Inspired by \cite{Sonneville_2014}, in \cite{Grazioso_SORO2019} a helical shape function defined with the exponential map is used to express the shape of a manipulator. It is given in terms of the pose at the base ($X=0$) and the strain $\bm{\epsilon}$.

\subsubsection{Relative modal-Ritz reduction}
\label{sec::RmodalRitz}
\noindent The modal Ritz approach has been recently applied to the strain fields of Cosserat rods in the context of the Geometric Variable-Strain (GVS) approach, \cite{Renda_RAL2020}, \cite{Boyer_TRO_2020}, \cite{Mathew_2021}. Equation \eqref{RitzGal} thus becomes $\bm{\epsilon}(X,t) = \bm{\Phi}(X) \bm{q}(t) $ (Fig. \ref{fig:Reductions}(c)). In this case, the Ritz coefficients of the strains stand for the generalized coordinates of a set of homogeneous transformations along the soft robot, similar to the joint transformations for rigid ones \cite{Murray}. This (relative) strain parametrization, produces a highly reduced set of ODEs in the classical form:
\begin{equation}
\label{general_dynamic_strain}
\bm{M}(\bm{q})\ddot{\bm{q}} + \bm{C}(\bm{q},\dot{\bm{q}})\dot{\bm{q}} + \bm{Q}_{int}(\bm{q}, \dot{\bm{q}})  = \bm{Q}_{ext}(\bm{q}, \dot{\bm{q}}),
\end{equation}
where $\bm{q}$ is then comparable to the vector of joint angles of a rigid robot. In \cite{Boyer_TRO_2020}, the GVS was successfully validated against the GE-FEM with standard benches in statics and dynamics. The results showed that the approach can provide very good results in terms of accuracy with a very few number of generalized coordinates. Moreover, in contrast to absolute Ritz methods, the strain functions do not need to fulfill any boundary conditions. 
However, the price to pay for this is that, unlike the usual absolute Ritz reduction (such as the FEM), it handles double space integrals which are not easy to calculate. Recent progress has been made to overcome this difficulty and obtain the reduced model (\ref{general_dynamic_strain}). In \cite{Mathew_2021}, D'Alembert's principle of virtual works is used to project the Cosserat model \eqref{diff_kin}, \eqref{SE(3)_Cosserat_equilibrium}, from the space of pose fields $\bm{g}$ onto that of strain coordinates $\bm{q}$. This approach has been achieved in a purely analytical way. It first exploits the fact that since equation \eqref{gprimo} is a system of homogeneous first order ODEs of matrix form  $\bm{Y}' = \bm{Y} \bm{A}(X)$, the exponential representation of its solutions can be obtained through the Magnus expansions, yielding, for a soft manipulator, to the geometric map:
\begin{equation}
\label{Magnus}
    \bm{g}(X) = \text{exp} \left( \widehat{\bm{\Omega}}(\bm{q},X) \right),
\end{equation}
where $\text{exp} $ represents the exponential map in $SE(3)$, and $\widehat{\bm{\Omega}}(\bm{q},X)$ is the Magnus expansion of $\widehat{\bm{\xi}}$, truncated at a desired order of approximation. Second, the strain reduction and \eqref{Magnus} are introduced into the expressions of $\dot{\bm{\eta}}$ and $\bm{\eta}$ obtained by analytical integration of the model of velocities and accelerations along the rods \eqref{transp}. This  provides the kinematic map $\bm{\eta}=\bm{J}(\bm{q})\dot{\bm{q}}$ and its time derivative $\dot{\bm{\eta}}=\bm{J}(\bm{q})\ddot{\bm{q}} + \dot{\bm{J}}(\bm{q})\dot{\bm{q}}$, with $\bm{J}(\bm{q})$ the Jacobian between the twist and $\dot{\bm{q}}$-spaces. Finally, applying this change of space in the contributions of external and inertial forces of the weak form of the virtual works, while those of the elastic and actuation ones are directly deduced of  the projection of the active constitutive law \eqref{active_constitutive_law} on the strain modes, we obtain the equations of motion in the classical Lagrangian form \eqref{general_dynamic_strain}. Alternatively, to this analytical Jacobian-based approach, in \cite{Boyer_TRO_2020} a numerical Newton-Euler based approach has been proposed to calculate \eqref{general_dynamic_strain}. In this case, the complex double space integrals required by the strain based parametrization are automatically performed through the two passes of the continuous computed torque algorithm of \cite{Boyer_TRO_2006}, whose inputs and outputs are reduced on the strain modes. In \cite{Renda_RAL2021}, the GVS approach is further extended to the modeling of concentric tube robot systems, including the modeling of the tube's insertion motion in quasi-steady conditions. In \cite{Boyer_2022}, the same approach is applied to the unsteady case to address the "paradoxical" dynamics of the sliding spaghetti \cite{Carrier1949}.
The GVS approach has been implemented in a Matlab Toolbox, \textit{SoRoSim}, for the simulation of soft, rigid and hybrid robots \cite{Mathew_2021}. A special case of the GVS is the Piecewise Constant-Strain approach (PCS) \cite{Renda_ICRA2018}, \cite{Renda_TRO2018}, where the strains in the sections are assumed to be constant, an approach which has been extended to closed-chain geometries in \cite{Armanini_TROFinRay}.
In \cite{Caasenbrood_SORO_2022}, a similar approach to PCS is presented. Based on the constant curvature assumption, it also incorporates hyperelastic and viscoelastic material behaviours. A planar restriction of the GVS, named Polynomial Curvature (PC) approach was introduced in \cite{DellaSantina_RAL2020} and \cite{DellaSantina_CDC2020} to address the control of planar soft manipulators. A similar approach was also used in \cite{Odhner_TRO2012} to model planar elastic flexure joints.

\subsection{Analytical-based resolutions}
\label{sec::EulerBernoulli}
\noindent In some specific loading conditions (especially when the rod is subject to concentrated external loads, and not distributed ones), some analytical solutions to equation \eqref{EB} can be obtained in terms of elliptic functions. To illustrate this, let us consider the case of a simply supported beam subjected to a concentrated compressive force $F$ applied at its tip. The analytical solution of the Euler Bernoulli equation then yields \cite{Bigoni}:
\begin{equation}
\label{EB_sol}
F = \frac{4 EI}{l^2} \left[\mathcal{K} \sin\left(\frac{\theta(0)}{2}\right) \right]^2 \; \text{,}
\end{equation}
where an uniform cross section along the length of the beam is assumed, and $\mathcal{K}$ is a complete elliptic integral of the first kind. It should be noted that equation \eqref{EB_sol} provides an implicit expression of the base angle $\theta(0)$ as a function of the applied load $F$. Once $\theta(0)$ calculated for a given $F$, the rotational field $\theta$ can be obtained through integration of \eqref{EB}, and used to get the displacement field of the rod in terms of other elliptic functions.
When applicable, this approach can offer great advantages for the study of some specific phenomena related to soft robots design.
EB rod theory was applied to statics and dynamics of a soft arm with a continuously rotating clamp subject to a tip dead load for the purpose of analysing snap-back phenomenon and post buckling behaviour \cite{Armanini2017}. In \cite{Simaan2009}, kineto-static modeling of multiple-backbone continuum robots is addressed with EB solutions able to express the deflected shapes of the backbones over a sub-segment of the robot. In the paper \cite{Xu_Simaan}, planar EB theory is applied to the modeling of continuous tendon-driven robots, and the static equilibrium of end and spacer disks is derived. In \cite{Polygerinos2015}, EB theory is applied to the modeling of a fiber-reinforced bending actuator. The soft actuator is modeled as a Neo-Hookean material, where the strain energy functional of \eqref{W_Odgen} reduces to $U = \frac{\mu}{2}(\lambda_1^2 + \lambda_2^2 + \lambda_3^2 - 3)$, with $\mu$, the shear modulus. The principal nominal stresses are deduced from the principal Cauchy stresses of \eqref{sigma_Odgen}. Considering the fiber reinforcement on the circumferential direction ($\lambda_2 =1$) and the material incompressibility, provides the three nominal stress and the Lagrange multiplier $p$ in terms of $\mu$ and $\lambda_1$. The bending moment is then deduced from stresses at the top and bottom layers, and set equal to that of the internal air pressure against the distal cap of the actuator.
The actuator force is obtained from the torque balance, assuming that the actuator is constrained in a flat configuration and that no internal moments is generated under pressurization. In \cite{Cacucciolo_2016}, the approach is extended in order to capture the effect of pressure on the lateral surface of the inner chamber of the actuator. In \cite{Olson2020}, a soft arm made of longitudinal pneumatic actuators is described in terms of its curvature with \eqref{EB}, a stretch field along the center line, and the bending plane defined by an angle at the base. The balance of internal and the external loads is then formulated, together with that of the cross section. The model is finally solved numerically, discretizing the arm into a finite number of sections and driving the residual of the static balance to zero.

\section{Geometrical Models}
\label{sec::Geom}
\noindent The main characteristic of the approaches in this Section is that they rely on the assumption that the deformed shape of the soft body resemble a specific geometrical shape. 
All these approaches are based on a representation of the soft body which falls in the definition of a Cosserat rod.
The main difference from the approaches described in Sec. \ref{sec::Director} is that, when moving from the kinematics to the static and dynamic equations, these methods do not rely on the PDEs \eqref{CR}, but built from the generalized coordinates specifically used to represent the body geometry. We can then conclude that, while the configuration of these models can vary, being discrete or continuous, they all have their roots in some set of shape coordinates mapped to the configuration by:
\begin{equation}
\label{Geom}
\bm{q} \in \mathbb{R}^n \mapsto \bm{g}(\bm{q})\in SE(3).
\end{equation}
The equilibrium is then given by the Euler-Lagrange equations or other equivalent principles of dynamics. Note that all these approaches can be considered as some approximations of the Cosserat model starting from the model of a sequence of deformable curves. As a result, depending on their material consistency, the resulting models can be equivalent to the ones derived in the previous section, with particular reference to the Ritz-based approach of Sec. \ref{sec::AmodalRitz} or \ref{sec::RmodalRitz}. Two main groups of approaches fall in this category: the \textit{functional} models and the Piecewise Constant-Curvature (PCC) models.

\subsection{Functional approaches}
\label{sec::Functional}
\noindent Functional approaches probably represent one of the first attempts to model soft robotic devices. Their main characteristic is that they all use a chosen mathematical function to describe the desired space curves representing the geometry of the robot. One of the first example falling in this category is developed in \cite{Hirose1994}, which employs a serpenoid curve to describe the kinematics of snake-like robots. Then, the kinematics of the robot is obtained by simple forces and torques equilibrium, considering the action of the ground on the robot and the internal force components \cite{Hirose2009}.
One other important example is the so-called modal approach, that have been originally presented for the modeling of hyper-redundant robots both in statics  \cite{Chirikjian_TRO_1994} and in dynamics \cite{Chirikjian1994}, while a survey of the approach is presented in \cite{Chirikjian2015}. The central concept of the approach is that of "\emph{backbone curve}", i.e. a 3D curve $\bm{r}$ parameterized by the integral representation $\bm{r}(X,t) = \int_{0}^{X} \lambda(s,t) \bm{t}(s,t) ds$, where $\lambda(X,t)$ represents a length scaling factor, while $\bm{t}(X,t)$ is the unit tangent vector, parameterized by any spherical kinematics representation (e.g. Euler angles, quaternions). In words, the deformed axis of the robot is represented as a curve growing along a direction field $\bm{t}(X,t)$ from the base to its tip, with a magnitude rate $\lambda(X,t)$.
In order to obtain the complete description of the robot geometry, the curve is equipped with a set of orthonormal frames and a roll distribution that describes the twist of the robot. At the end, the robot geometry is defined by a reduced set of shape functions, describing the backbone curve itself and a model of the twist. In \cite{Chirikjian_TRO_1994}, a set of four independent shape functions $S_i(X,t) = \{\lambda(X,t) \; \theta(X,t) \; \phi(X,t) \; \psi(X,t) \}$ is used, in which $\theta(X,t)$ and $\phi(X,t)$ are the two angles defining $\bm{t}(X,t)$ and $\psi(X,t)$ is the roll distribution function. The inverse kinematic problem then consists in finding the set of shape functions satisfying the task constraints, which are usually represented by the end-effector positioning. For this purpose, a modal approach is employed and each $S_i(X,t)$ takes the form $S_i(X,t)=\sum_{j=1}^{N_{S_i}} \Phi_{ij}(X)q_{ij}(t)$, where $q_{ij}$ are modal participation factors and the $\Phi_{ij}(X)$s, $N_{S_i}$ modal functions, which are chosen by the robot programmer to fulfill the task constraints. At the end, the robot shape is fully described by the modal factors $q_{ij}$, which represent the generalized coordinates of the system $\bm{q}$ in \eqref{Geom}. In \cite{Webster2010} it has been proved that, when the curvature of the body is assumed to be constant, this approach provides a transformation which is equivalent to \eqref{TW}, presented in the following Section. In one other work by the same authors \cite{Chirikjian2015}, variational approaches are presented in order to compute the optimal curve shapes that comply with both joint and task constraints. The same kinematic model was also employed in \cite{Ivanescu2007} and extended to the dynamical modeling of coiling continuum robots.
Another possible function that can be used to represent the robot geometry consists in the Pythagorean hodograph (PH) curves \cite{Singh2018}. The backbone is represented using quadratic polynomials, which are functions of the chosen control points, representing the generalized coordinates of the system. In order to obtain the optimal quadratic polynomials to compute the control points, the minimizing of the potential energy is applied. Finally, a neural network model is employed to predict the effects of loads on the position of the robot, including the case of variable loads. In \cite{Wiese2019} the kinematics of a soft pneumatic actuator is modelled representing its backbone with cubic Hermite splines (also called cspline). 
In this way, the backbone is defined by two control points and two control orientations (vectors). The control points are obtained experimentally and an optimization procedure is carried out to fit the cspline with the given backbone. To complete the description of the robot configuration, the orientation along the curve is finally obtained assuming the minimizing of the torsion of the actuator's backbone. In \cite{Rao_RAL_2022} the backbone of tendon driven continuum robots is represented as a sequence of Euler arc splines, which are directed arcs whose curvature varies in arithmetic progression. This paper represent an example of a modeling technique that is based on the Constant Curvature assumption, as the models presented in Sec. \ref{sec::PCC}, but employs a functional representation of the robot's kinematics. In \cite{Manaswi2019}, elliptic Fourier descriptors are employed to describe soft deformable morphologies. Compared to the other methods described in this Section, this approach does not model the soft robot through its backbone, but uses a closed curve to represent the boundary of the 2-dimensional regions that it occupies. This is obtained thanks to elliptic Fourier descriptor, through a procedure that fits a closed curve to a set of 2-dimensional points with arbitrary precision. In particular, the image of the region occupied by the robot is extracted from experimental recordings and its contour is identified through a discrete representation. Finally, the four coefficients of the Fourier series are computed, providing the description of the shape.

\subsection{Constant Curvature Models} \label{sec::PCC}
\noindent Constant curvature is often viewed as a desirable characteristic in continuum robots, due to the simplifications it enables in kinematic modeling as well as in real-time control and other useful computations. This is motivated by the fact that actuators with a path parallel to the mid-line on a cylindrical manipulator produce a constant curvature shape, in the absence of external forces. For these reasons, the constant-curvature assumption has been successfully applied in a great number of continuum robots modeling approaches. In these models, a soft body is represented by a finite number of circular arcs, each having a curvature that is constant in space. For these approaches, the coordinates $\bm{q}$ in \eqref{Geom} are specifically obtained to describe the circular arcs geometry.
We can distinguish two main groups of PCC approaches: the \textit{Kinematics-based models} are developed from a kinematical relation between the actuator and the arcs parameters, while the \textit{Mechanics-based models} are based on the mechanical description of the problem. Finally, a survey on some of the main PCC approaches can be found in \cite{Webster2010} and \cite{Chawla2018}.

\subsubsection{Kinematics-based models}
\noindent Once the continuous body is represented as a finite set of CC segments, each of these can be represented by a finite set of arc parameters and it is possible to obtain a map from them to the task space of the robot. Different parameters can be used to describe a CC segment, yielding to different kinematics maps. One of the most popular sets of arc parameters that have been proposed consists of triplets of curvature $\kappa$, the angle of the plane containing the arc $\phi$, and arc length $l$, which define $\bm{q}$ in (\ref{Geom}). Different approaches have been proposed to obtain the kinematic map from these arc parameters. In \cite{Webster2010} it has been proved that they all provide an identical transformation from the arc base to any point $s \in [0,l]$ of the robot arm (noting $\cos=\text{c}$ and $\sin=\text{s}$):
\begin{equation}\label{TW}
\bm{T}(\kappa, \phi, s) =
\begin{bmatrix}
\text{c}\phi \text{c} \kappa s & -\text{s}\phi & \text{c}\phi \text{s}\kappa s & \frac{1}{k} \text{c}\phi(1 - \text{c}\kappa s)\\
\text{s}\phi \text{c} \kappa s & \text{c}\phi & \text{s}\phi \text{s}\kappa s & \frac{1}{k} \text{s}\phi(1 - \text{c}\kappa s)\\
-\text{s}\kappa s & 0 & \text{c}\kappa s & \frac{1}{\kappa} \text{s}\kappa s \\
0 & 0 & 0 & 1
\end{bmatrix}.
\end{equation}
One way to obtain \eqref{TW} is based on the use of Denavit-Hartenberg (D-H) parameters, \cite{Gravagne2003}, \cite{Jones_TRO_2006}, \cite{Walker2013}. 

\noindent In \cite{Gravagne2003}, under the constant curvature assumptions, the continuous backbone of the robot is fitted with a virtual conventional rigid-link manipulator and modified D-H parameters are obtained to consider the coupling imposed by the curvature in a continuum system, providing the standard homogenous transformation matrix \eqref{TW}. In a following work \cite{Jones_TRO_2006}, a similar, improved, approach is presented, finally providing the transformation from three parallel actuating tendons to the arc parameters:
\begin{equation}
\label{Transf_Tendon}
    \begin{array}{l}
l = \dfrac{l_1 + l_2 + l_3}{3} \; ,\phi = \tan^{-1} \left(\dfrac{\sqrt{3}}{3} \dfrac{l_2 + l_3 -2l_1}{l_2 - l_3} \right) \; , \vspace{0.03cm}\\
\kappa = \frac{2}{d} \dfrac{\sqrt{l_1^2 +l_2^2 + l_3^2 -l_1l_2- l_1l_3 -l_2l_3}}{l_1+l_2+l_3}
    \end{array}
\end{equation}
where $(l_1,l_2,l_3)^{T}$ are actuator's lengths, while $d$ is the distance from the center of the section to the actuator. This approach has been applied for the modeling of a great number of continuum robots \cite{Escande2015}, \cite{Lakhal2016} and \cite{Mahl2014}, \cite{Falkenhahn2015}. 

\noindent One of the main restrictions of these PCC models is that the used parametrization and kinematic maps can implicitly provide a numerical singularity that occurs when the curvature tends to vanish ($\kappa\rightarrow 0$), resulting in an infinite or undefined radius of curvature. In order to overcome this limitation, different solutions have been introduced. In \cite{Godage_IROS2011} and \cite{Godage_BB2015} the rotational and position components of the homogeneous transformation \eqref{TW} are represented in modal form, similarly to the models presented in Sec. \ref{sec::Functional}. In particular, the entries in \eqref{TW} are numerically approximated using multivariate Taylor series expansions for actuator's lengths variables at $\bm{0}$.
Based on this formulation, a Lagrangian approach was developed in \cite{Godage_ROBIO2011} and \cite{Godage_ICRA2011} for the spatial dynamics of a single section continuum arm and this was further extended to multi-section arms in \cite{Godage_IJRR2016}. Other approaches have been mostly focused on the definition of a different parametrization and transformation map to describe the geometry of the soft bodies. One alternative is provided by the exponential map of \cite{Webster_IJRR2006}, \cite{Webster_TRO2009} and \cite{Sears2006}. In particular, \eqref{TW} can be obtained using the exponential coordinates from the Lie Group theory seen in Sec. \ref{sec::EnergeticA}. Considering that the transformation for a circular arc is the composition of a rotation $\bm{\zeta}_{rot}$ with an in-plane $\bm{\zeta}_{inp}$ transformation, one has:
\begin{equation}\label{exp}
    \bm{T}(\kappa, \phi, s ) = \text{exp}(\hat{\bm{\zeta}}_{rot}\phi) \text{exp}(\hat{\bm{\zeta}}_{inp}(\kappa) s).
\end{equation}
In \cite{Webster_TRO2009}, this approach is applied to the modeling of concentric tubes robot. It is assumed that the two tubes have the same stiffness and are torsionally rigid, while applying torques to one another. Given the arc assumption, these torques can be considered to be uniform. The in-plane bending model is obtained employing EB linear equation and the resultant curvature of two overlapping tubes is obtained through a force balance in analogy with linear springs connected in parallel. 
Finally, the mapping from the arc-parameters to the Cartesian poses of the tube cross-sections is obtained with \eqref{exp}. Similarly, in \cite{Allen2020}, the arcs are expressed with an axis-rotation parametrization. The origin of the axis $\bm{\omega}$ is positioned at point $\bm{\rho}$ on the horizontal plane at the base of the section and $\bm{\omega}$ is perpendicular to $\bm{\rho}$. $\|\bm{\rho}\|$ is the radius of arc curvature, while $\|\bm{\omega}\|=l/\|\bm{\rho}\|$ is equal to the angle between the proximal and distal positions when $\bm{\rho}$ is used as the vertex of the angle. The CC arc is then parameterize by $(\bm{\rho},l)$ or equivalently $(\bm{\omega},l)$ which defines $\bm{q}$. The forward kinematics can thus be obtained through the exponential maps. 
Straightforward geometry finally provides the mapping between the configuration parameters and the length of the actuating tendons \eqref{Transf_Tendon}. In essence, this is a screw theory method, since $\bm{\omega}$ is exactly the axis of the screw motion required to cover the arc bend, as shown in \cite{Renda2017}.

\noindent One other example of a parametrization that does not entail the singularity in the proximity of a null curvature configuration is provided in \cite{DellaSantina2020}, where for each CC section composing the robot, the configuration of a segment is defined as linear combinations of four arcs included in the section's volume. In particular, given the difference in length between opposite arcs, $\Delta_{x,i}$ and $\Delta_{y,i}$, through geometrical considerations, it is possible to obtain a transition map from the ''\textit{conventional}'' arc parameters $(\kappa, \phi, l)$ and the new parametrization $(\Delta_x, \Delta_y, \delta l)$, where $\delta l$ represents the change in the length of the section with respect to the at rest position. Algebraic steps are finally used to obtain the explicit expression of the analogous of \eqref{TW}, in terms of the new parameters.

\subsubsection{Mechanics-based models}
\noindent Some authors have sought to draw mechanical consequences, by applying static laws to a soft arm whose shape is described by the PCC. In \cite{Camarillo_TRO2008} both the forward and the inverse kinematics for a tendon driven manipulator with parallel routines are derived in this way. More precisely, considering a single tendon that can experience only a constant tension $ u $ along its length, force and moment balances are used to define the internal reaction forces acting on a section of the beam. This model is then extended to consider the presence of a finite number of tendons, based on an analogy with a system of linear springs acting in parallel. In the considered planar case, the generalized coordinates $\bm{q}$ describing the configuration of the arm are the curvature $\kappa$ and an axial stretch measurement $\epsilon_a$. The static equilibrium leads to the description of the mechanical response of the manipulator in the form $\bm{K} \bm{q} = \bm{B}^T \bm{u}$, where $\bm{B}^T$ represents the tendon moment arms and the axially directed tangent. Relating the tendon tension and displacements is then possible to obtain the mechanical response of the tendons in terms of their displacements $\bm{y}$, $\bm{y} = \bm{C}_m \bm{u}$, where $\bm{C}_m$ is a compliance matrix, which is function of $\bm{B}^T$ and of the manipulator and tendon stiffness matrices and lengths. Finally, this yields to
\begin{equation}
\label{eq:Camarillo_final}
    \bm{y} = \bm{A}^{\dag} \bm{q}
\end{equation}
where $\bm{A} = \bm{K}^{-1} \bm{B}^T \bm{C}_m^{-1}$ is the forward kinematics transformation matrix. Equation \eqref{eq:Camarillo_final} provides a mechanics-based relationship between the beam and the tendon configuration, analogous to transformations \eqref{Transf_Tendon}. In a following work \cite{Camarillo_TRO2009}, the 3D static model is also developed.

\noindent In \cite{Sadati_Frontiers2017} the statics of a braided pneumatic continuum manipulator is addressed. In this case, the effect of cross-sectional deformations on the arm deformation is studied using a strain energy function based on the Cauchy-Green stretch. The study shows that the PCC parametrization fails while releasing the PCC constraints gives better results.  Another source of mechanical inconsistency in the PCC is related to the modeling of torsion which is absent from the model mainly concerned with bending. In order to solve this issue, several developments of the PCC were proposed. In \cite{Rone_TRO_2014} and \cite{Rone2014}, the virtual power method is used to obtain the TDCRs robot dynamics. The considered robot geometry is that of \cite{Xu_Simaan}, i.e., an elastic backbone connected by rigid disks and three tendon cables. The disks are considered rigid, while the backbone and the cables sub-segment are treated as circular arcs. The relative pose of two adjacent discs is obtained from the two constant orthogonal curvatures $(\beta,\gamma)$ and an additional lumped torsional angle $\epsilon$ of the subsegment, which define the $\bm{q}$-vector.

\noindent The torsion model has become particularly critical in continuum robotics of CTR, where it is at origin of the so called "snapping" phenomenon. In \cite{Dupont2010}, the authors provide first the design conditions necessary for the validity of PCC approach, and conclude with a general model for CTRs. In this further model, the relative twist angle between two tubes is defined as a function of the arc length and then used to write the moment equilibrium and to impose the compatibility equations, enforcing the coincidence of the tube centerlines. The torsional strain is then deduced from the Cosserat rod equilibrium \eqref{CR}. The resulting BVP was then solved by exploring the entire input-output set. Based on this model, the stability analysis was performed in \cite{Xu2014} and \cite{Hendrick2015}.


\section{Discrete Models}
\label{sec::Discr}
\noindent In this section we discuss the approaches where the configuration of the system is discrete from the very beginning (i.e. the system is not discretized at the resolution level or through some assumptions on its backbone's geometry). In particular, we can distinguish three main groups: the lumped-mass, the pseudo-rigid, and the discrete rods models.

\subsection{Lumped-mass Models}
\label{sec::Lumped}
\begin{figure}[ht]
\centering
\includegraphics[width=0.4\columnwidth]{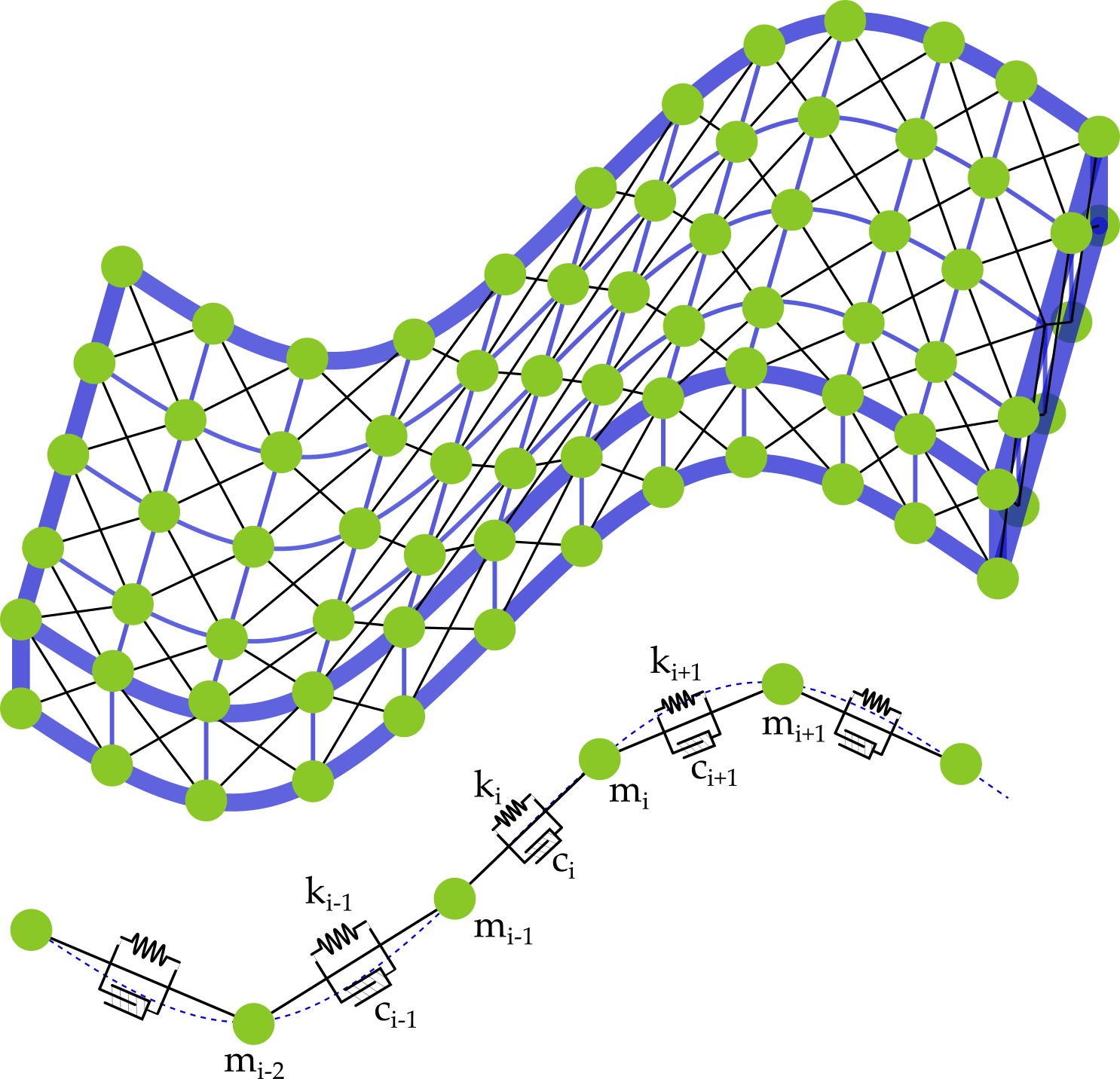}
\caption{\footnotesize The lumped mass representation of a continuous rod}
\label{fig:Lumped}
\vspace{-0.2cm}
\end{figure}

\noindent One of the simplest approaches to modeling a continuous body is to represent it as a set of discrete masses, dampers and springs, Fig. \ref{fig:Lumped}.
In this way, the governing equations can be obtained by energy methods or applying classical Newton equations to describe the propagation of the forces from one particle to the other in the configuration space:
\begin{equation}
Q \subset \mathbb{R}^3 \times \mathbb{R}^3 \times \mathbb{R}^3 \dots \times \mathbb{R}^3.
\end{equation}
The main advantage of lumped-parameter mechanics models is that they are characterized by a simple structure, easy to adapt to the capture of complex phenomena such as nonlinear friction, material hysteresis, and inertial dynamics. On the other end, to reach the same fidelity of the continuum mechanics models or the FEM approach, they require a high number of DOFs and a data-expensive system identification procedure.

\noindent In \cite{Habibi2019}, mass-spring-damper arrays are used to model the large deformations of a continuum surface (LDCS) actuated by continuum arms. In particular, the 3D problem of a system composed of two interconnected layers of LDCSs is considered, where bending and shear deformations are allowed. The surface is modeled as a lattice where the masses are defined as the nodes of the system and they are connected through linear spring and dampers. Assuming a uniformly stiff surface, the elastic constant of the springs and the damping coefficients are assumed to be constant. Considering an isotropic linear elastic material and assuming a straight undeformed configuration, the equations of motion for each mass is obtained by combining direct application of Newton's second law and Hooke's law.
In \cite{Jung2011}, a discrete lumped-mass approach is presented for the modeling of tendon driven medical robotic catheters. The axial and the bending stiffness of the model are adjusted through the selection of the spring stiffness and their radial location, while the smooth motion of the catheter is modeled through the dampers.
Because of their simplicity, lumped mass models are particularly suited for large robotics simulations library, such as the one presented in \cite{Austin_ICRA2020}. Titan is a GPU-accelerated C++ software library for the modeling of soft bodies and multi-agent robots at massive scales. The main benefit of this approach is the possibility to perform parallel computing, resulting in good computational performances.

\subsection{Pseudo-rigid Models}
\label{sec::PseudoRigid}
\noindent The opportunity to exploit well known and established rigid robotics theories for soft robots motivated the family of models that are described here. The soft bodies are represented as series of rigid links which are connected by revolute, universal or spherical joints, as seen in Fig. \ref{fig:Reductions}(d). Thus, in general, the configuration space of the soft body is given in the form :
\begin{equation}\label{SE(3)_configuration}
Q \subset SE(3) \times SE(3) \times SE(3) \dots \times SE(3).
\end{equation}
While these approaches can provide satisfactory results for the modeling of hyper-redundant or snake-like robots, when dealing with continuous elastic structures they provide a low order of spatial approximation accuracy, in addition to the expensive identification procedure as before. In \cite{Kutzer2011}, a dexterous catheter manipulator is represented as a series of pin joints connected by rigid links. In \cite{Khoshnam2013}, a pseudo-rigid 3D approach is applied for the modeling of a steerable ablation catheter. The catheter is treated as a cantilever beam, which is represented as four rigid links connected by three revolute joints and three torsional springs. \cite{Venkiteswaran2019} presents a 6 DOF pseudo-rigid model for continuum manipulators subject to multiple external loads. In this case, a flexible manipulator is represented as four rigid links of given length which are connected by three joints having 2DOF each. In \cite{Yu2005}, the dynamical modeling of compliant mechanism is addressed and a flexible beam is represented as a mass-less rigid body with a torsion spring attached at one end.
In \cite{Mochiyama2002}, the modeling of hyper-flexible manipulators is addressed using a serial rigid chain, where the number of kinematic DOF goes to infinity. The backbone curve is first described as a continuous curve with extended Frenet frames attached to each point along its length. Then, for the numerical simulations, the backbone curve is approximated by a serial chain of rigid bodies.
In \cite{Ganji2009}, a steering catheter is represented as a combination of three sections: the virtual base of the distal shaft, represented as a prismatic joint; the bending section of the distal shaft, represented as two revolute joints, one prismatic joint and two revolute joints; the distal end of the catheter, which is treated as a rigid body. Similarly to the PCC parametrization of \cite{Gravagne2003}, D-H parameters are then employed to obtain the forward kinematics of the model. In \cite{Sfakiotakis_2014} a pseudo-rigid approach is proposed for modeling an octopus inspired 8-armed swimmer. In particular, each arm is modeled as a kinematic chain of cylindrical rigid segments which are connected by planar rotatory joints. The first joints in the chain, connecting the arms of the octopus to the main body, are modeled as actuated rotatory joints, while the other joints are modeled as (unactuated) rotatory linear spring and damper elements.

\subsection{Discrete rods}
\label{sec::DiscreteRods}
\noindent The computer graphics research community represents an important source of inspiration for soft robotics modeling. This is the case of the approaches that are described in this Section, which are all based, with different extent, on the discrete elastic rod (DER) approach, originally introduced in \cite{Bergou_2008}, \cite{Bergou_2010} and \cite{Audoly_2018}. In this formulation, the inextensible Kirchhoff rod assumptions \eqref{KIRCH} are considered and, similarly to the Cosserat rod model, the configuration $\bm{g}$ of an elastic rod is given by a curve $\bm{r}(X)$, representing the centerline, and a material frame, where the first axis is always tangent to the curve. The Bishop (natural) frame $\{\bm{t}, \bm{v}, \bm{w} \}$ is then introduced, providing, for a given centerline, the most \textit{relaxed} frame, i.e with zero twist. Finally, parallel transport allows to define the evolution of the Bishop frame along the mid-line.
\begin{figure}[h]
\centering
\includegraphics[width=0.6\columnwidth]{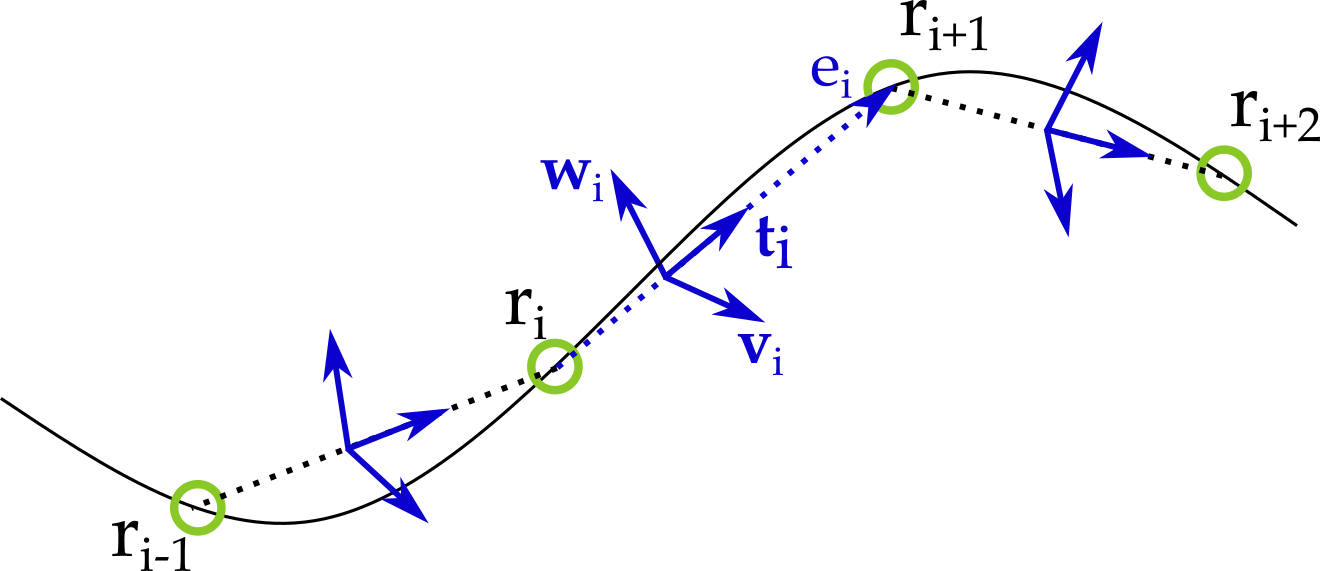}
\caption{\footnotesize Representation of a rod in the DER approach}
\label{fig:DER}
\vspace{-0.2cm}
\end{figure}
In this way, the model provides a simple parametrization of the cross-sectional poses, given by the curve $\bm{r}(X)$ and an angle $\theta(X)$ between the Bishop and the cross-sectional frame. The rod is then discretized into a series of finite nodes (or vertices) connected by straight segments (or edges). Each node is characterized by a position vector $\bm{r}_i$, while each segment is associated to the edge vector $\bm{e}_i = \bm{r}_{i+1} - \bm{r}_i$ and its tangent unit vector $\bm{t}_i$, Fig. \ref{fig:DER}. In this way, the discrete curvature associated with the $i$th vertex takes the form:
\begin{equation}
    \kappa_i = \frac{2 \sin\phi_i}{1 + \cos\phi_i} = 2\tan\frac{\phi_i}{2} \; \text{,}
\end{equation}
where $\phi_i$ represents the turning angle between two consecutive edges. Similarly, it is possible to define the discrete curvature bi-normal. Each $i$-th vertex is also characterized by a total mass $M_i$, which is the average mass of the edges meeting in the vertex, while the mass moment of inertia for each edge can be obtained through volume integration. Finally, a discrete bending energy and a discrete twist energy can be obtained in terms of the generalized coordinates $(\bm{r}_i, \theta_i, \phi_i)$.

\noindent Following the DER formulation, in \cite{Gazzola2018} the three dimensional space of a soft filament is represented by a set of vertices $\bm{r}_i(t)$ and a set of material frames $\bm{R}_i(t)$. Each vertex is characterized by a linear velocity, a concentrated mass and a set of concentrated external forces. Extension, shear and axial deformations are considered. Finally, through the spatial integration, the discrete governing equations are obtained and they are solved using a symplectic, second-order scheme. Similarly to other approaches, the external physical interactions are included in the external forces and moments vectors, while the internal physical effects (such as those modeling a muscular activity) are added to the internal forces and moments resultants. In a recent work \cite{Gazzola2021}, this model has been implemented in an open-source simulation environment, \textit{Elastica}, for the modeling and simulation of the dynamics of slender rods.

\noindent One other example of a soft robotics modeling technique following the DER approach is presented in \cite{Goldberg2019}. In particular, the planar case of the discrete elastic rod formulation is considered, where the rod is free to move on a plane, while the torsional deformations are neglected. Moreover, the notion of folding for a straight rod is introduced to apply the approach to tree-like architecture, which are frequently encountered in the soft robotics community.
In \cite{Huang2020}, the DER-based formulation is further enhanced to include frictional contact, inelastic collisions and inertial effects. In particular, considering a planar case, a Rayleigh damping matrix is defined to formulate the internal damping forces vector which is added to the external forces. In order to model the contact and the friction between the robot and an unstructured ground, whose normal direction can vary with the horizontal axis, a Coulomb law is employed. In \cite{Megaro_2017}, the DER is hybridized with a pseudo-rigid body model is used to represent compliant mechanisms. Based on this model, the authors provided a computational design tool to perform different optimization tasks related to soft robotics.

\section{Surrogate Models}
\label{sec::DataDriven}
\noindent With respect to the models presented so far, a complete different approach to tackle the modeling of a soft robot consists in using large sets of data that are derived from various forms of external sources. These approaches are often referred to as \textit{surrogate} or \textit{data-driven} models and a survey on some of them is presented in \cite{Daekyum2021} and \cite{Wang_2021}. While one of the main benefits of these solutions is that they do not require a physical model, on the other hand they rely on large amounts of representative data, that are sometimes difficult to collect. Most of data-driven approaches presented for soft robotics modeling involve neural networks models.

\subsection{Neural Networks}
\label{sec::NN}
\begin{figure}[h]
\centering
\includegraphics[width=0.5\columnwidth]{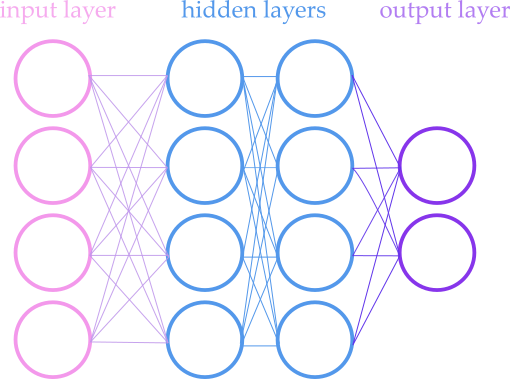}
\caption{\footnotesize Scheme of a neural network}
\label{fig:NN}
\vspace{-0.4cm}
\end{figure}

\noindent Neural networks (NN) have been proved to be an effective tool to solve many kinds of non linear problems in different application fields, including robotics. As the name suggests, they are inspired from the biological neural networks that operates animal brains: the artificial neurons, which represent the elementary units of NNs, can transmit a signals to the other neurons. The signal is usually a real number and the output of each neuron is computed as a non-linear combination of all the inputs. Neurons are connected by edges and they are all characterized by weights adjusted during the learning process. Usually, neurons are gathered in layers: the first layer is also called \textit{input layer}, the last one is called \textit{output layer} and the intermediate layers are called \textit{hidden layers}, Fig. \ref{fig:NN}.
NN results in a high dimensional set of nested functions:
\begin{equation}
\label{NN}
\bm{y} = f_M(\bm{A}_M, \dots f_2(\bm{A}_2,f_1(\bm{A}_1,\bm{x}))\dots),
\end{equation}
where $\bm{x}$ are the values of the input nodes, $\bm{A}_i$ are the edges weights, $ f_i $ are the activation functions and $\bm{y}$ are the values of the output nodes. In most soft robotics applications, the input and the output layers represent the actuation variables $\bm{u}$ and the shape parameters $\bm{q}$. The learning process results in an optimization over the network weights, which is performed through back-propagation, a form of chain rule, where, after each forward step through the network, a backward pass is performed to compute the network Jacobian and to adjusts the model's weights accordingly. Neural networks have been used to approximate the kinematics, the statics and also the dynamics description of soft robots. Typically, Feed-forward neural network (FNN) are employed for kinematics and static modeling.
On the other hand, Recurrent Neural Networks (RNN) are usually preferred for dynamical modeling.

\noindent The first examples of the employment of NN algorithms for the control of continuum robots were presented in \cite{Braganza2007}. In particular, a controller for continuum robots uses a FNN component to compensate the dynamic uncertainties of the system, in an attempt to reduce the uncertainty bound. In \cite{Melingui2014} the forward kinematic modeling of a bionic assistant trunk is obtained through Multilayer Perceptron (MLP) and Radial Basis Function (RBF) neural networks, which are both class of FNNs. The input neurons propagate the input variables to the following layers, while each neuron in the hidden layer is associated with a RBF kernel (usually Gaussian). The data are obtained experimentally, measuring the arm's tip position at varying the actuation pressure, using a trilateration algorithm. In one other paper by the same authors \cite{Melingui_TRO_2015}, the same handling assistant trunk is modeled, both with a data-driven and a pseudo-rigid modeling approach. With regards to the latter, the trunk is modelled as a series of rigid vertebrae connected by a total of four prismatic joints. For the data-driven approach, a modified Elman neural network is employed.
FNNs are also employed in \cite{Giorelli_TRO_2015}  and \cite{Giorelli_BB2015}, which deal with the inverse \textit{kinetics} of a cable driven soft manipulator. While the \textit{kinematical} model describes the configuration of the robot without considering the involved loads, the \textit{kinetics} model relates the motion of the robot with its actuation forces. The direct kinetics model is obtained using a geometrically-exact model \cite{Renda_BB2012}. Thus, a FNN, taking the tip position as an input and giving the cable tension as an output, is employed. An experimental data collection phase is carried out, using a set of cable tensions and obtaining the tip position with an infrared vision system. These data were used for the optimization and training of the FNN. After the training phase, the performance of the FNN are measured on test sets, using the output of the FNN as the input of the direct kinetics model.
One other example of a machine learning formulation for the global inverse kinematics of continuum manipulators was presented in \cite{Thuruthel_SORO2017}. The data samples are generated by continuous motor babbling and a single hidden-layer artificial neural network is employed to learn directly the mapping $(\bm{x}_{i+1}, \bm{q}_i) \mapsto \bm{q}_{i+1}$, where $\bm{x}$ is the pose of the end effector. In \cite{Thuruthel_BB2017} a dynamic model for open-loop control of soft robotics manipulators is presented. The PCS model described in Sec. \ref{sec::RmodalRitz} is employed to obtain the dynamics of a cable driven robot operating underwater. Considering the case where the robot and the task space have the same number of DOF, the forward dynamics of the system \eqref{FEM_dynamic} is formulated using only the task space variables and the direct mapping between the states of the task space variables and the control inputs can be obtained. Once the training is completed, an open loop predictive controller is developed through a trajectory optimization that is carried out with an iterative sequential quadratic programming. In \cite{Thuruthel_TRO2019}, the closed-loop controller is also implemented.
The inverse kinematic modeling of a bionic trunk is addressed in \cite{Rolf2014}. In particular, a learning phase is carried out considering a volume of desired cartesian position for the robot's tip, defining a finite set of target vertices. The inverse model is asked to estimate a posture that allows to move the effector to each vertex, and the training is carried out until the distance between the target and the actual position for each vertex is minimized.

\subsection{Data-driven Order Reduction}
\label{sec::LOR}
\noindent Some modeling tools employ different forms of data-driven order reduction to efficiently approximate the physical model. In \cite{Bruder2021}, a Koopman Operator Theory is employed for a data driven controller of soft robots. A dynamical system is represented in a infinite function space $\mathcal{F}$, which is composed of real-valued functions inside the state of the system domain \cite{Hassan2017}. The elements $q \in \mathcal{F}$ are called observables. The Koopman operators, denoted by $\mathcal{U}_t$, is defined as the  linear transformation $\mathcal{U}_t q = q \circ T_t$,
where $\circ$ indicates the composition operation, while $T_t$ is the flow (or dynamic) map of the system. In other words, the Koopman operator lifts the dynamics of the system from the state space to the space of the observables, describing the evolution of the observables $q$ along the trajectories of the system. Its main advantage is that it provides a linear representation of the flow of a non-linear system, but in the infinite-dimensional space of the observables. The discrete approximate of the Koopman operator can be obtained from a set of experimentally measured state, given in the form of snapshot pairs.
Some approaches have been focused on the model reduction of FEM that were presented in Sec. \ref{sec::FEM}. Model reduction methods are based on the projection of the FE equations of motion to attractive sub-spaces of smaller dimensions. In this way, the size and the computational time of the simulation are drastically lowered, allowing the application of FE methods for control purposes. In \cite{Duriez_TRO2018}, a snapshot proper orthogonal decomposition (POD) is employed to generate relevant bases to be used for the order reduction. More into details, the solution $\bm{q}$ of equation \eqref{eq_Sofa} is expressed as a truncated expansion of orthonormal vectors, which depend on the constraints $\bm{\lambda}$. The orthonormal basis is then set to minimize the sum of all the errors that are generated by the projection of the exact solution onto the basis. In \cite{Thieffry_2019} this approach is further extended for the development of a low-order controller and observer, while in \cite{Goury_RAL2021} it is applied for the reduction of self-collision contact forces. One other example of a FE order reduction is presented in \cite{Chenevier2018} for modeling soft robots made of hyperelastic materials and actuated by cables or tendons, with a special focus on contact problems.

\section{Software Implementation}
\label{sec::Comput}
An important aspect of the above models and their numerical resolution, is related to their software implementation. There are several industrial as well as open-source soft robotic software/toolboxes available. Here we briefly summarize the most popular ones. In Sec. \ref{sec::FEM} we introduced Sofa, one of the earliest open-source platforms for physics-based simulation. Its SoftRobots plugin employs a multi-modal description of a problem, allowing the presence of several representations (mechanical, thermal and visual) of the same object. This multi-modal representation allows the simulation of scenarios involving the interaction of different components (rigid and/or soft bodies, fluids).
PyElastica is the Python implementation of Elastica \cite{Gazzola2021}, described in Sec. \ref{sec::DiscreteRods}. SoRoSim \cite{Mathew_2021} is a MATLAB toolbox that uses the geometrically exact PVS model described in Sec. \ref{sec::RmodalRitz}. One other example of a MATLAB  toolbox is TMTDyn which employs discretized lumped systems with reduced-order models \cite{Sadati_IJRR_2021}. SoMo (Soft Motion) \cite{Graule_IROS2021} couples the pseudo-rigid model discretization of Sec. \ref{sec::PseudoRigid} with a rigid body physics engine of Python to model soft robots. ChainQueen \cite{Hu2019} and IPC-Sim (Incremental Potential Contact Simulator) \cite{Minchen2020} are simulators oriented towards computer graphics rather than mechanical systems. ChainQueen is a Python toolbox that uses the Moving Least Square Material Point Method, a hybrid Eulerian/Lagrangian FEM which uses both particles and grids to simulate soft bodies. IPC-Sim solves extreme non-linear volumetric elastodynamic models using FEM. Finally, some toolboxes are developed for specific applications, such as the DiffAqua \cite{Pingchuan2021}, an optimization toolbox for soft underwater swimming robots.

\section{Critical analysis and view points}
\label{sec::Discussion}

\begin{table*}[b]
\centering
\tiny
\begin{tabular}{ 
    |m{1.7cm}       
    |m{0.1cm}       
    |m{1.7cm}       
    |m{1.5cm}       
    |m{5.3cm}       
    |m{5.2cm}       
    |@{}m{0pt}@{}}  
 \hline

    \multicolumn{2}{|c|}{\textbf{THEO. BACKGROUND}} & 
    \centering \textbf{REDUCTION / DISCRETIZATION} & 
    \centering \textbf{GEN. COORDINATES} &
    \centering \textbf{DESIGNS} &
    \centering \textbf{USES}  
    & \\[8pt] \hline

 \multirow{6}{1.7cm}[-5em]{\textbf{CONTINUUM MECHANICS MODELS}: the discretization and/or reduction is applied at the numerical resolution level.} 
  
    & \rotatebox{90}{3D}  
    & FEM, ANCS-FEM, MLS-FEM
    & \multirow{2}{1.5cm}[-1em]{\centering configurational fields values at the nodes} 
    & They can be applied to a \textbf{vast number of problems}, including 2D and 3D designs and actuation forces involving 3D problems (ballooning actuators and electroactive polymers), which are modeled as constraints. On the other hand, they do not appear as the optimal solution for systems which can be modeled by ad-hoc reduced theories.
    &  Thanks to commercial software adapted to 3D multiphysics contexts, they can be used \textbf{to design new soft robot concepts}. They are computationally expensive and provide models with a huge number of dofs \textbf{not adapted to control and fast simulation}. Sofa is today the optimal trade-off between accuracy and computational efficiency for 3D soft robotics. They require additional reduction treatments, to be used for control purposes.
    & 
    \\[40pt] \cline{2-3} \cline{5-6}
  
    & \multirow{5}{0.1cm}[-5em]{\rotatebox{90}{Directors}} 
  
    & ABS. NODAL-RITZ: GE-FEM
    & 
    & \multirow{5}{5.3cm}[-1em]{They provide the most efficient techniques for \textbf{1D problems mostly characterized by bending deformations}, i.e. systems that can be described as (combination of) beams and that do not involve large cross-section inflation \textbf{Thread-like actuators actions} are simply added as external or, sometimes, internal wrenches. On the other hand, they struggle in the modeling of other forms of actuation that involve 3D problems. Relative parametrizations such as the \textbf{GVS approach} can be easily extended to hybrid soft-rigid systems. When the circular arcs assumption holds, they reduce to \textbf{mechanically consistent and singularity free PCC}.}  
    &  \multirow{5}{5.2cm}[-3em]{GE-FEMs are fast and accurate for systems made of beams and rigid bodies, but are difficult to implement. In principle, other directors approaches are as accurate as GE-FEM. Energetic approaches based on modal  Ritz reduction, provide \textbf{highly reduced models} in explicit standard forms,  \textbf{well suited for control}. Absolute variables are more difficult to reduce than relative ones. Non-energetic approaches based on \textbf{shooting resolution}, are tailored  for\textbf{ real-time simulation}.}
    & 
    \\[10pt] \cline{3-4}   
  
    & 
    & REL. NODAL-RITZ 
    & \centering strain field values at the nodes  
    & 
    & 
    &  
    \\[10pt] \cline{3-4} 

    &  
    & EULER BERNOULLI 
    & \centering bending angle function
    & 
    & 
    &  
    \\[10pt] \cline{3-4} 
    
    &  
    & NON ENERGETIC 
    & \centering not available
    & 
    & 
    &  
    \\[10pt] \cline{3-4} 
    
    &  
    & MODAL RITZ 
    & \multirow{2}{1.5cm}[-0.5em]{\centering spatial coefficients of strain or configurational fields} 
    & 
    & 
    &  
    \\[10pt] \cline{1-3} \cline{5-6}
    
 \multicolumn{2}{|l|}{\multirow{2}{1.8cm}[0em]{\textbf{GEOMETRICAL}: the backbone is represented by a specific geometrical shape.}} 

    & FUNCTIONAL 
    & 
    & \multirow{2}{5.3cm}[0.5em]{PCC approaches are a specialization of GVS. They are only valid when \textbf{the circular arcs assumption holds}, i.e. in specific conditions of external and actuation loading. Similarly, functional approaches are built around a reference curve and they struggle in modeling situations which divert from this assumption.}
    & \multirow{2}{5cm}[0em]{When applicable, they are \textbf{ideally tailored to kinematic control} and offer explicit analytic maps between actuation and configuration spaces. Functional approaches often struggle to  model torsion in a consistent manner.}
    &  
    \\[12pt] \cline{3-4}  
    
    \multicolumn{2}{|l|}{} 
    & PCC 
    & \centering arc parameters or tendon lengths
    & 
    & 
    & \\[12pt] \hline

  \multicolumn{2}{|l|}{\multirow{3}{2.5cm}[-4em]{\textbf{DISCRETE}: the material model is a-priori discretized}} 
  
    & LUMPED MASS 
    & \centering  mass positions  
    &  \multirow{2}{5.3cm}[1em]{These are applications of traditional rigid theories and, for this reason, they are particularly simple and useful when applied to \textbf{hybrids systems}, combining soft and rigid components of arbitrary geometries. On the other hand, they struggle in the modeling of aspects that are more specific for soft robotics applications, such as the constitutive equations or the modeling of distributed actuation.} 
     & \multirow{2}{5.2cm}[0em]{They are conceptually simple to implement. However, like FEM, they handle a large number of discrete dofs, \textbf{requiring reduction techniques to produce models suitable for control}. Unlike FEM, they suffer from an \textbf{inconsistent discretization}, especially for internal strain energy.}
    & 
    \\[15pt] \cline{3-4}
  
    \multicolumn{2}{|l|}{} 
    & PSEUDO-RIGID 
    & \centering joint coordinates  
    & 
    & 
    \\[15pt] \cline{3-4}  \cline{5-6}
  
    \multicolumn{2}{|l|}{}  
    & DISCRETE RODS 
    & \centering configurational fields values at the edges
    & Their applicability is comparable to that of absolute nodal Ritz methods  based on directors' approaches, such as GE-FEM.  They are well suited  for beam and rigid body systems. They are easier to implement than FEMs, but still need to be adapted to soft robots.
    & Unlike other discrete approaches, they rely on a consistent discretization of the Cosserat model, based on discrete differential geometry. They have been developed for \textbf{fast interactive simulation} in computer graphics and should be \textbf{adapted to real-time simulation and control} of soft robotic.
    & 
    \\[30pt] \cline{1-6}  

 \multicolumn{2}{|l|}{\multirow{2}{2.5cm}[-0.5em]{\textbf{SURROGATE}:  the model is obtained by regression of a large sets of data.}}

    & NEURAL NETWORKS 
    & \centering not applicable
    & \multirow{2}{5.3cm}[-1em]{They are \textbf{physics-agnostic} approaches that can be possibly applied to any design and scenario. However, the resulting models are not generalizable to new designs and conditions.} 
    & \multirow{2}{5.2cm}[-1em]{They generate \textbf{dedicated algorithms} that can be used for \textbf{fast simulation and control}.}
    &  
    \\[10pt] \cline{3-4} 
    
    \multicolumn{2}{|l|}{} 
    & ORDER REDUCTION 
    & \centering depend on the reduction space basis 
    & 
    & 
    \\[10pt] \hline
 \end{tabular}
\caption{ \footnotesize Summary of the main modeling families, their fundamental properties and their design applications.}
 \label{tab::summ}
 \vspace{-0.3cm}
\end{table*}



\noindent Providing realistic models able to be operated computationally in real time or faster, to solve direct and inverse problems of the wide variety of soft robots designed so far is an unattainable dream.
Therefore, the choice of a modeling technique is the result of a compromise between realism (accuracy) and computational efficiency, which is largely dictated by the user's specific needs. For example, designers may be more interested in realism in order to capture the concept of a new actuator in a first phase, and in a second phase by computational efficiency of forward (kinematics, statics, dynamics) algorithms that can help optimization techniques requiring many simulations with different sets of design parameters. In the same way, closed-loop control strategies can allow some relaxation of accuracy in favor of designing inverse algorithms (kinematics, statics, dynamics) capable of running online. Based on this trade-off, here follows a critical analysis of the modelling approaches presented above.\\
\noindent Surrogate models of section VIII have the great advantage of being conceptually applicable to all systems, without physical modeling effort, and can provide fast dedicated algorithms compatible with real-time control. However, for each robot design, a huge database of measurements has to be built to feed the learning process.  
\noindent As regards physics-based models, the only approaches that are physically consistent at all levels of modelling, (geometry, kinematics and dynamics), are those based on continuum mechanics of sections III, IV and V, as well as the discrete rods method of section VII.C. Potentially, they can be used to solve both direct problems (simulation) and inverse problems (control). Like data-driven models of section VIII, commercial FEM software have the advantage of being applicable to complex systems of arbitrary 3D geometry in multi-physics contexts.
For these reasons, they are well suited to the design of new actuation principles or to the study of specific complex phenomena. However, they are often expensive in terms of computation time and poorly adapted to robotics specific control problems. To speed up the computations, the direct material discretizations in Section VII.A,B, may be appealing at first glance, since they allow any continuous system to be replaced by a discrete one with lumped parameters. However, although they can provide mechanically consistent discrete formulations suited to fast simulation, they require complex identification methods to relate the lumped parameters to the distributed original ones. In this regard, it is worth mentioning that it is one of the historical motivations of the FEM to have overcame this "naive" discretization. With the same goals in mind, Sofa has made great efforts in recent years to revisit and optimize 3D FEM for user-friendly modeling and rapid simulation of soft robots interacting with complex environments. However, due to the large number of nodal dof, all these approaches remain difficult to exploit for control purposes, and one of the challenges they face is to develop efficient reduction numerical methods compatible with the highly nonlinear character of soft robots (see section VIII.B). When possible, a natural reduction is to consider special geometries such as shells and rods. While the former remain little represented in robotics, the latter are widely used to model multibody systems composed of rods and rigid bodies. In FEM, the use of beam-specific finite elements then becomes natural, and the GE-FEM based on Cosserat beams of section III.B, seems to be a good choice, as it is very efficient both in terms of accuracy and computational cost. However, it still suffers from a lack of existing commercial software dedicated to robotics and is difficult to implement from scratch. \\
\noindent Based on the same models (based on 1D directors), the other methods in Section IV inherently enjoy similar accuracy to that of GE-FEM, and represent reliable alternatives for soft robots that can be described as beams (or a combination of beams). This includes robots with concentric tubes, continuous parallel robots (CPRs), or soft arms that do not involve large cross-sectional inflation, but also systems with thread-like actuators, such as TDCRs. Moreover, unlike FEM, they were specifically developed for robotic purposes, whether for real-time simulation based on the shooting method, or for the design of model-based control laws. In the latter case, modal Ritz approaches as the GVS can provide highly reduced dynamic models in the usual explicit Lagrangian form of rigid robots, and thus facilitate the transfer of methods from rigid to soft robotics.\\ 
\noindent On the other hand, although computationally very efficient and ideally suited for kinematic control, the geometric approaches of VI.B. are based on certain kinematic simplifications that are not consistent with continuum mechanics. In particular, PCC remains mechanically consistent only for specific actuation designs and when gravity is negligible. Note that although widely used, this approach being a sub-case of the fully consistent GVS, it should be advantageously replaced by this alternative approach. Regarding the other simplified models of Section VI, while functional approaches built around a reference curve, are conceptually natural and can be computationally efficient, they often struggle to capture the torsional state in a consistent way. Also note that in contrast to modelling approaches based on a direct material discretization, the discrete rod method is fully consistent since it is based from the beginning on the discrete differential geometry of the Cosserat theory (and from that point of view, should be classified into the directors based approaches of section III.B). Moreover, it has been developed from the start for fast (interactive) simulation by the computer graphics community and from this point of view, it represents a promising perspective for soft robotics. Finally, Table 1 summarizes the main modeling families in relation to their uses and robot designs.

\section{Conclusions} \label{sec::Conclusions}
\noindent Soft robotics was developed in the last two decades and, legitimately, the main efforts so far have been focuses on the simulation of the robot's static and dynamic behaviour. Now the efforts of many researchers are mostly towards the extension of these approaches for control \cite{DellaSantina_Review2021}, \cite{Renda_RAL2022} and for their integration with optimization tools \cite{Moritz_2021}, \cite{Armanini_RAL2022}. Once more, the main sources of inspiration towards the achievements of these goals are more likely to come from other research disciplines. Indeed, while going through the papers that have been presented so far for the modeling of soft robots, we tackled a great number of disciplines that represent their theoretical foundations: continuum and solid mechanics, computational mechanics, machine learning, computer graphics, just to name a few. This interdisciplinary is probably what makes soft robotics so interesting, attracting scientists from different research fields. On the other side, being the topic so widespread, it is not easy to grasp it and this constitutes the main motivation behind this manuscript. In this way, we were able to recognize more clearly the uniqueness and the commonalities between the different techniques that have been presented so far, in the effort to untangle such a vast research topic.
\section*{Acknowledgement}
\noindent This work was supported by the US Office of Naval Research Global under Grant N62909-21-1-2033 and in part by Khalifa University of Science and Technology under Grants CIRA-2020-074, RC1-2018-KUCARS. This work was also partly supported by the French National Research Agency (ANR) through the COSSEROOTS research project ANR-20-CE33-0001 (2020-2024).

\bibliographystyle{ieeetr}
\bibliography{Review_BIB}

\end{document}